\documentclass[10pt,twocolumn,letterpaper]{article}

\usepackage{iccv}
\usepackage{times}
\usepackage{epsfig}
\usepackage{graphicx}
\usepackage{tabularx}
\usepackage{amsmath}
\usepackage{amssymb}

\usepackage{xcolor}
\usepackage{array, makecell}
\usepackage{multirow}
\usepackage{float}
\usepackage{gensymb}
\restylefloat{table}
\restylefloat{figure}

\definecolor{DarkGreen}{rgb}{0.0, 0.2, 0.0}

\definecolor{DarkBlue}{rgb}{0.0, 0.0, 0.8}

\newcommand{\ul}[1]{\underline{#1}}

\usepackage[breaklinks=true,bookmarks=false]{hyperref}

\iccvfinalcopy 

\def\iccvPaperID{9} 

\usepackage[utf8]{inputenc}
\usepackage[T1]{fontenc}
\begin{document}

\title{SharpNet: Fast and Accurate Recovery of Occluding Contours\\ in 
Monocular Depth Estimation}

\author{Micha\"el Ramamonjisoa$^1$ $\quad \quad$ Vincent Lepetit$^1$\\
	\hspace{0em} $~^1$LIGM (UMR 8049), École des Ponts, UPE\\
	{\hspace{-0em} $\texttt{\small 
			\{michael.ramamonjisoa,vincent.lepetit\}@enpc.fr}$}
}

\maketitle

\begin{abstract}

We introduce SharpNet, a method that predicts an accurate depth map given a 
single input color image, with a particular attention to the reconstruction of 
occluding contours: Occluding contours are an important cue for object 
recognition, and for realistic integration of virtual objects in Augmented 
Reality, but they are also notoriously difficult to reconstruct accurately. For 
example, they are a challenge for stereo-based reconstruction methods, as 
points around an occluding contour are only visible in one of the two views. 
Inspired by recent methods that introduce normal estimation to improve depth 
prediction, we introduce novel terms to constrain normals, depth and occluding 
contours predictions. Since ground truth depth is difficult to obtain with 
pixel-perfect accuracy along  occluding contours, we use synthetic images for 
training, followed by fine-tuning on real data. We demonstrate our approach on 
the challenging NYUv2-Depth dataset, and show that our method outperforms the 
state-of-the-art along occluding contours, while performing on par with the 
best recent methods for the rest of the images. Its accuracy along the 
occluding contours is actually better than the ``ground truth'' acquired by a 
depth camera based on structured light. We show this by introducing a new 
benchmark based on NYUv2-Depth for evaluating occluding contours in monocular 
reconstruction, which is our second contribution.
\end{abstract}

\section{Introduction}

Monocular depth  estimation is a  very ill-posed  yet highly desirable  task for
applications such as  robotics, augmented or mixed  reality, autonomous driving,
and scene understanding in general. Recently, many methods have been proposed to
solve this problem using Deep Learning approaches, either relying on supervised
learning~\cite{Eigen14, Eigen2015PredictingDS, Laina2016DeeperDP, FuCVPR18-DORN}
or on self-supervised learning~\cite{monodepth17,XieDeep3D,PoggiTrinocular},  
and  these methods often yield very impressive results.

		
\begin{figure}
\begin{center}	
	\resizebox{\linewidth}{!}{
	\begin{tabular}{>{\centering\arraybackslash}m{1.3cm}
					>{\centering\arraybackslash}m{0.34\linewidth}
					>{\centering\arraybackslash}m{0.34\linewidth}}				
							
		Jiao \textit{et al.}~\cite{Jiao2018LookDI} &
		\includegraphics[width=0.9\linewidth]{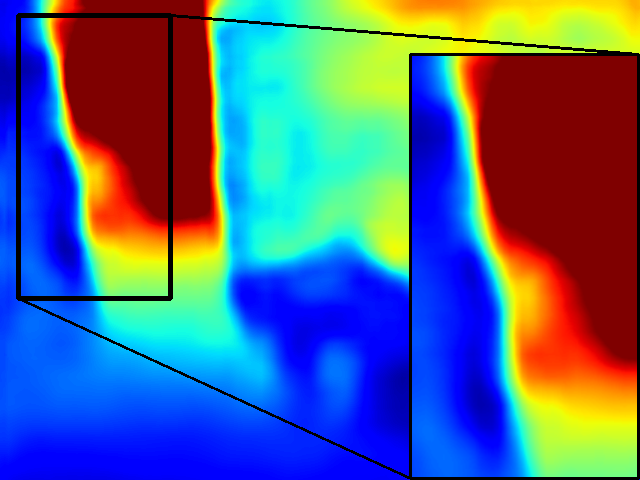}
		& 
		\includegraphics[width=0.9\linewidth]{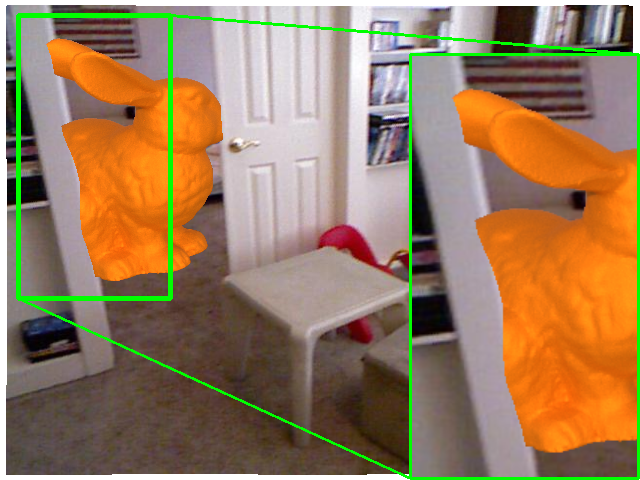}
		\\
		NYUv2-Depth Ground Truth Depth&
		\includegraphics[width=0.9\linewidth]{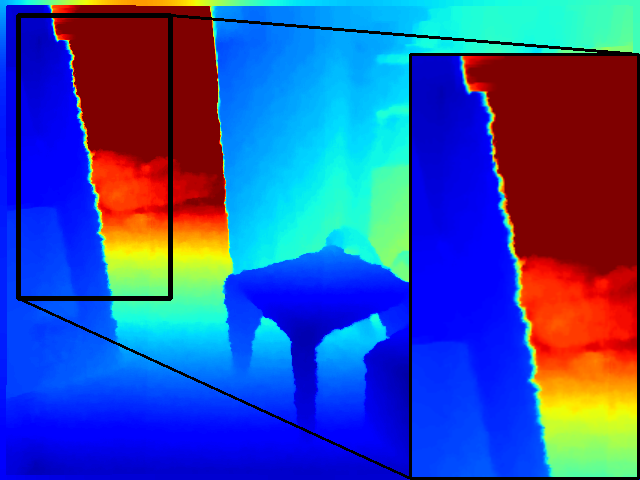}
		& 
		\includegraphics[width=0.9\linewidth]{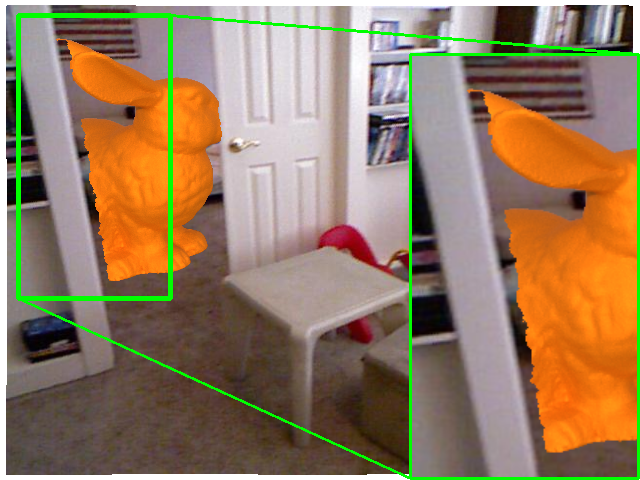}
		\\		
		Ours &
		\includegraphics[width=0.9\linewidth]{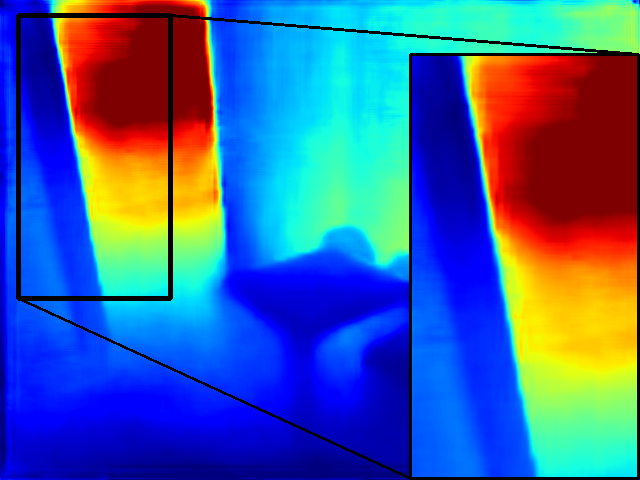}
		& 
		\includegraphics[width=0.9\linewidth]{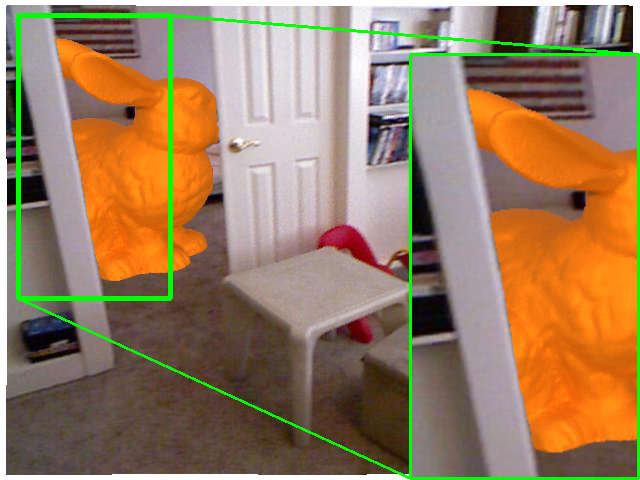}
		\\
		Manual insertion &
		\includegraphics[width=0.9\linewidth]{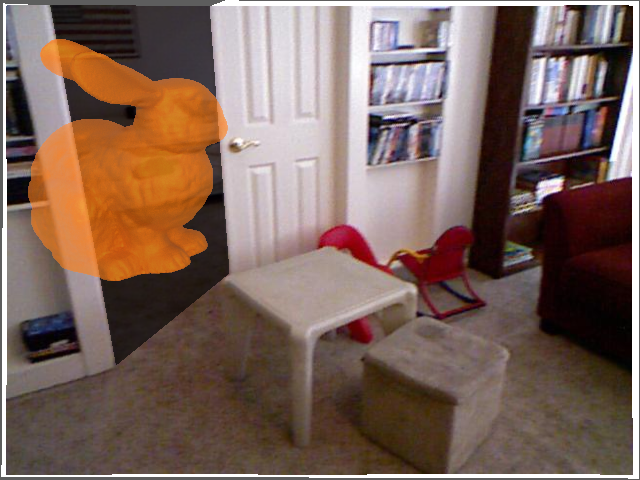}
		&
		\includegraphics[width=0.9\linewidth]{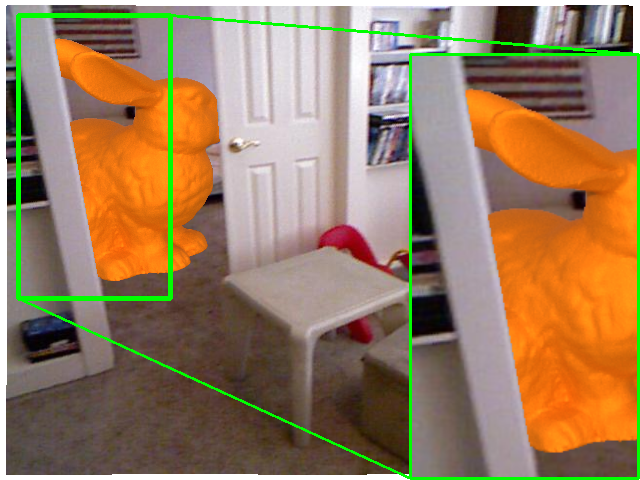}\\
	\end{tabular}}	
\label{fig:one}
\caption{Our SharpNet method shows significant  improvement over 
state-of-the-art monocular depth estimation methods in  terms  of  occluding  
contours  accuracy, and even produces sharper edges along these contours than 
structured-light depth cameras.
In this figure we augment an RGB image from NYUv2~\cite{Nyuv2} with a virtual 
Stanford  
rabbit using different depth maps for  occlusion-aware integration. The first 
three rows show the depth map used for occlusion-aware insertion (left)  and 
resulting augmentation (right).  
An error of only a few pixels along occluding contours can significantly degrade
the realism of the integration, comparatively to manual insertion (last row) 
using a binary mask.
\vspace{-0.2cm}
}
\end{center}
\end{figure}

Despite recent advances in monocular depth estimation, occluding contours 
remain difficult to reconstruct correctly from depth as shown in 
Fig.~\ref{fig:one}, while they are still an important cue for object 
recognition, and for augmented reality or path planning, for example.  
This has several causes: First, the  depth annotations  of training  
images are  likely to  be inaccurate along the  occluding contours, if the 
depth  annotations are obtained with a stereo reconstruction method or a 
structured light  camera.  This is for example the case for the NYUv2-Depth 
dataset~\cite{Nyuv2}, which is an important benchmark used by many  recent 
works for evaluation. This is because on one or both sides of the occluding 
contours lie 3D points that are visible in only one image, which challenges the 
3D reconstruction~\cite{Szeliski11}. Structured light cameras essentially rely  
on stereo reconstruction, where one  image is replaced by a known 
pattern~\cite{Han13},  and therefore suffer  from the  same problem. Secondly, 
occluding contours, despite their  importance, represent a small part of the 
images, and may not influence the loss function used during training if they 
are not handled with special care.

In  this paper,  we  show that  it  is  possible to  learn to reconstruct 
occluding contours more accurately by adding a simple term that constrains the 
depth predictions together with the occluding contours during learning.  This 
approach is inspired  by recent works  that predict the depths  and normals 
for  an input image, and enforce constraints between 
them~\cite{WangSurgeNIPS16,Qi2018GeoNetGN,Yang2018lego}.  A similar constraint 
between depth and occluding contours can be introduced, and we show that 
this results in better reconstructions  along the  occluding  contours,  
without degrading  the accuracy of the rest of the reconstruction.

Specifically, we  train a  network to  predict depths,  normals, and  occluding
contours  for an  input image,  by minimizing  a loss function that integrates
constraints between the depths and the  occluding contours, and also between 
depths and normals. We show that  these two constraints can be integrated in
a very similar way with simple terms  in the loss function.  At run-time, we can
predict  only  the  depth  values,  making our  method suitable for real-time 
applications since it runs at 150 fps on $640\times 480$ images.

We show  that each aspect of  our training procedure improves  the depth output.
In  particular, our  experiments show  that  the constraint  between depths  and
occluding contours is  important, and that the improvement is not only due to 
multi-task learning.
Learning to  predict the normals in addition to  the depths and the occluding  
contours  helps  the  convergence  of  training  towards  good  depth
predictions.

We demonstrate our  approach on the NYUv2-Depth dataset, in  order to compare it
against  previous methods. Since we need training data with pixel perfect depth 
annotation along the occluding  contours, we use synthetic
images to pretrain the  network  before fine-tuning on NYUv2-Depth. We simply 
use the object instance boundaries given by the synthetic dataset as training 
annotations of the occluding contours.  However, we only use the  depth ground  
truth as  training data  when finetuning  on the  NYUv2-Depth dataset.

A proper  evaluation of  the accuracy  of the  occluding contours  is difficult.
Since  the  ``ground truth''  depth  data  is  typically noisy  along  occluding
contours, as in NYUv2-Depth, an evaluation based on this data would not be 
representative of the actual quality.  Even with better depth data, identifying 
occluding contours automatically as ground truth depth discontinuities would be 
sensitive to the parameters used by the identification method~\cite{Canny} 
(see Fig.~\ref{fig:multi_canny}).

We therefore decided to annotate manually the occluding contours  in a subset 
of 100 images randomly sampled from the NYUv2-Depth validation set, which we 
call the NYUv2-OC dataset. Our annotations and our code for the 
evaluation of the occluding contours are publicly  available for comparison. 
We evaluate our method on this data in terms of 2D localization, in addition 
to evaluating depth estimation on the NYUv2-Depth validation set using more  
standard depth estimation 
metrics~\cite{Eigen14,Eigen2015PredictingDS,Laina2016DeeperDP}.  Our experiments
show  that  while  achieving  competitive results on the NYUv2-Depth  benchmark 
by  placing  second on  all of  them,  we outperform  all previous methods in 
terms of  occluding contours 2D localization, especially the current leading 
method on monocular depth estimation~\cite{Jiao2018LookDI}.

\section{Related Work}

Monocular depth  estimation from images made  significant progress  recently. 
In the following, we mainly discuss the most recent methods and popular 
techniques  that help monocular  depth estimation: Learning from  
synthetic data and using normals for learning to predict depths.

\subsection{Supervised and Self-Supervised Monocular Depth Estimation}

With   the   development   of large  datasets  of   images   annotated   with   
depth data~\cite{Nyuv2,Geiger2013IJRR,Sun-rgbd,Zhang2016pbrs,Matterport3D, 
McCormacSceneNetRGBD}, many supervised methods have been   proposed. Eigen 
\textit{et al.}~\cite{Eigen14,Eigen2015PredictingDS} used multi-scale depth 
estimation to capture global and local information to help depth prediction. 
Given the remarkable performances they achieved on both popular benchmarks 
NYUv2-Depth~\cite{Nyuv2} and KITTI~\cite{Geiger2013IJRR}, more work extended 
this multi-scale approach~\cite{LiICCV17,xu2017MS-CRF}.   Previous work also   
considered ordinal depth classification~\cite{FuCVPR18-DORN}     
or pair-wise depth-map comparisons~\cite{Cao2018MonocularDE} to add local and 
non-local  constraints. Our approach  relies on a simpler  mono-scale 
architecture,  making it efficient at run-time. Our constraints between depths, 
normals, and occluding contours guide learning towards good depth prediction 
for the whole image.

Laina \textit{et al.}~\cite{Laina2016DeeperDP} exploited the power of deep 
residual neural networks~\cite{He2016ResNet} and showed that using 
the more appropriate BerHu~\cite{BerHuOwen,BerHuZwald} reconstruction loss 
yields better performances. However, their end results are quite smooth around 
occluding contours, making their method inappropriate for realistic 
occlusion-aware augmented reality. 

Jiao \textit{et al.}~\cite{Jiao2018LookDI} noticed that the depth distribution 
of  the NYUv2 dataset is heavy-tailed. The authors therefore proposed an 
attention-driven  loss for the  network supervision,  and  paired the depth  
estimation  task with  semantic segmentation  to  improve performances  on  
the  dataset.  However, while they currently achieve the  best  performance 
on  the  NYUv2-Depth  dataset,  their approach suffers from a bias towards 
high-depth areas such as windows, corridors or mirrors. While this translates 
into a significant decrease of the final error, it also produces blurry    
depth maps, as one can see in Fig.~\ref{fig:one}.  By contrast, our 
reconstructions tend to be much sharper along the occluding boundaries as 
desired, and our method is much faster, making it suitable for real-time 
applications. 

Self-supervised learning methods have also become popular for monocular 
reconstruction, as they exploit the consistency between multiple 
views~\cite{monodepth17,XieDeep3D,PoggiTrinocular,Yang2018UnsupervisedLO, 
Yin2018UnsupervisedGeoNet, TengOcclusionAware2018}.  While such approach is
very appealing, it does not yet reach the accuracy of supervised methods in
general, and it should be preferred only when no annotated data are available 
for supervised learning.

\subsection{Edge- and Occlusion-Aware Depth Estimation}

Wang \textit{et al.}~\cite{WangSurgeNIPS16} introduced their SURGE method to 
improve scene reconstruction on planar and edge regions by learning to jointly 
predict depth and normal maps, as well as edges and planar regions. They then 
refine the depth prediction by solving an optimization problem using a Dense 
Conditional Random Field~(DCRF). While their method yields appealing 
reconstruction results on planar regions, it still underperforms 
state-of-the-art methods on global metrics, and the use of DCRF makes it 
unsuited for real-time applications. Furthermore, SURGE~\cite{WangSurgeNIPS16} 
is evaluated on the reconstruction quality around edges using standard depth 
error metrics, but not on the 2D localization of their occluding contours.

Many self-supervised methods~\cite{Yang2018UnsupervisedLO,Yang2018lego, 
Yin2018UnsupervisedGeoNet, TengOcclusionAware2018, monodepth17} have 
incorporated edge- or occlusion-aware geometry constraints which exist when 
working with stereo pairs or sequences of images as provided in the very 
popular KITTI depth estimation benchmark~\cite{Geiger2013IJRR}. However, 
although these methods can perform monocular depth estimation at test time, 
they require multiple calibrated views at training time. They are 
therefore unable to work on monocular RGB-D datasets such as 
NYUv2-Depth~\cite{Nyuv2} or SUN-RGBD~\cite{Sun-rgbd}.

\cite{Wang2016LFCameras, Jiang2018DepthEW} worked on occlusion-aware depth 
estimation to improve reconstruction for augmented reality applications. While 
achieving spectacular results, they however require one or multiple light-field 
images, which are more costly to obtain than ubiquitous RGB images.

Conscious of the lack of evaluation metrics and benchmarks for quality of edge 
and planes reconstruction from monocular depth estimates, Koch \textit{et 
al.}~\cite{Koch2018EvaluationOC} introduced the iBims-v1 dataset, a high 
quality benchmark of 100 RGB images with their associated depth map. This work 
tackles the low quality of depth maps of other RGB-D datasets such as 
\cite{Sun-rgbd} and \cite{Nyuv2}, and introduces annotations and metrics for 
occluding contours and planarity of planar regions. Our evaluation method of 
occluding contours reconstruction quality is based on their work.

\section{Method}

\newcommand{\bhC}{\widehat{\boldsymbol{C}}}
\newcommand{\hCi}{\widehat{C}_i}
\newcommand{\bhD}{\widehat{\boldsymbol{D}}}
\newcommand{\hDi}{\widehat{D}_i}
\newcommand{\bhN}{\widehat{\boldsymbol{N}}}
\newcommand{\bhNi}{\widehat{\boldsymbol{N}}_i}

\newcommand{\bC}{\boldsymbol{C}}
\newcommand{\bD}{\boldsymbol{D}}
\newcommand{\bI}{\boldsymbol{I}}
\newcommand{\bN}{\boldsymbol{N}}
\newcommand{\bM}{\boldsymbol{M}}
\newcommand{\bX}{\boldsymbol{X}}
\newcommand{\bn}{\boldsymbol{n}}
\newcommand{\bu}{\boldsymbol{u}}

\newcommand{\calL}{\mathcal{L}}

\newcommand{\normals}{\text{n}}
\newcommand{\depth}{\text{d}}
\newcommand{\boundaries}{\text{c}}
\newcommand{\superv}{\text{superv}}
\newcommand{\dc}{\text{dc}}
\newcommand{\dn}{\text{dn}}
\newcommand{\BerHu}{\text{BerHu}}
\newcommand{\Huber}{\text{Huber}}
\newcommand{\AL}{\text{AL}}

\newcommand{\IR}{\mathbb{R}}

\begin{figure}[H]
	\begin{center}
			\includegraphics[width=1\linewidth, height=4cm]{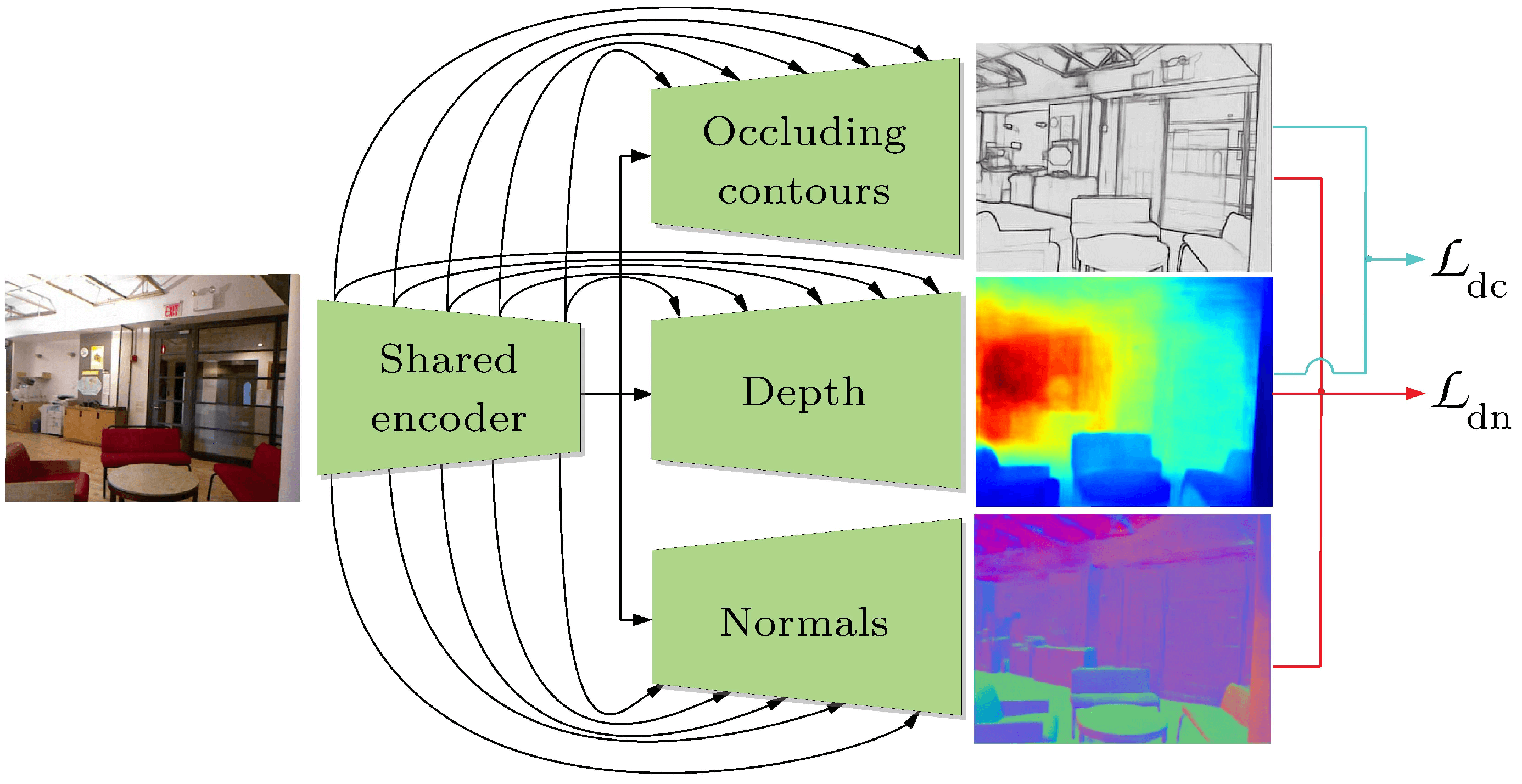}
	\end{center}
	\caption{The architecture of our ``U-net''-shape \cite{Unet} multi-task 
	encoder-decoder network. We use a single ResNet50 encoder which learns an 
	intermediate representation that is shared by all decoders. With this 
	setting, the representation generalizes better for all tasks. 
		We use skip connections between features of the encoder and of the decoder at corresponding scales.
		}
	\label{fig:architecture}
\end{figure}

\begin{figure*}[h]
	\begin{center}
		\begin{tabular}{>{\centering}m{0.045\linewidth}
				>{\centering\arraybackslash}m{0.127\linewidth}
				>{\centering\arraybackslash}m{0.127\linewidth}
				>{\centering\arraybackslash}m{0.127\linewidth}
				>{\centering\arraybackslash}m{0.127\linewidth}
				>{\centering\arraybackslash}m{0.127\linewidth}
				>{\centering\arraybackslash}m{0.127\linewidth}}
			~ & 
			$\quad$ RGB Input & $\quad$Depth GT & $\quad$Depth Pred & 
			$\quad$Normals 
			GT & $\quad$Normals Pred & $\quad$Contours Pred\\
			NYUv2 \cite{Nyuv2} &
			\includegraphics[width=1.2\linewidth]{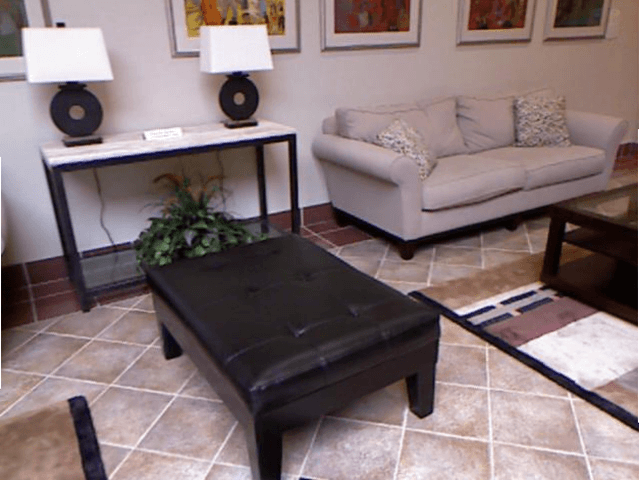} &
			\includegraphics[width=1.2\linewidth]{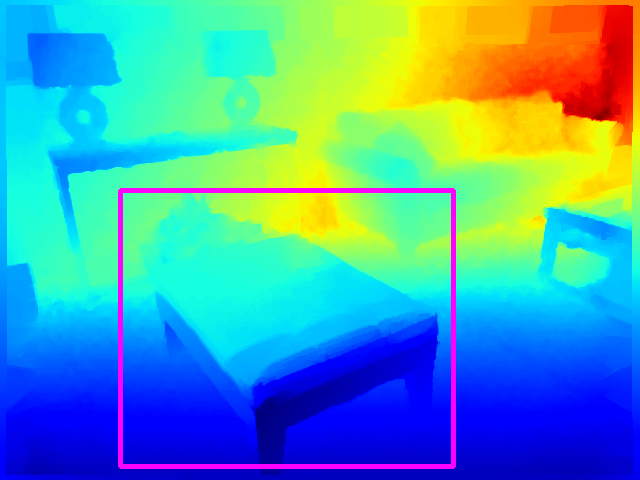}
			 &
			\includegraphics[width=1.2\linewidth]{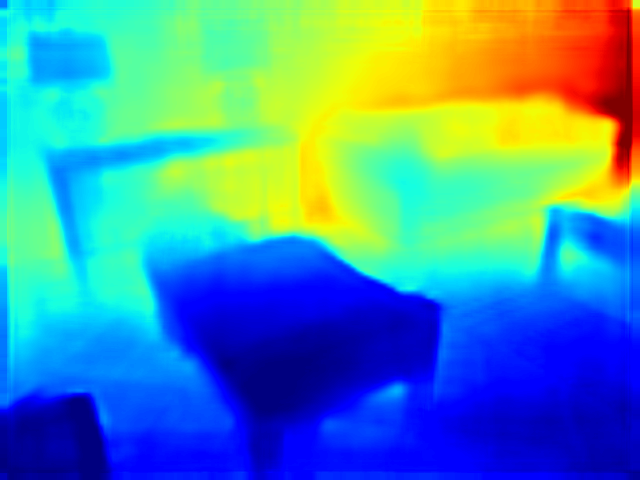} &
			\includegraphics[width=1.2\linewidth]{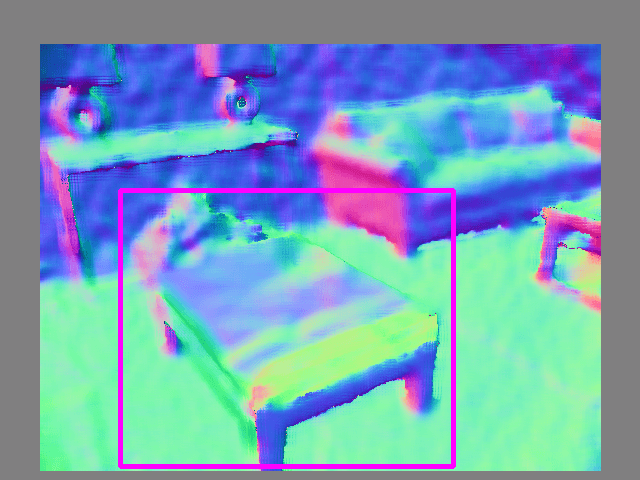}
			 &
			\includegraphics[width=1.2\linewidth]{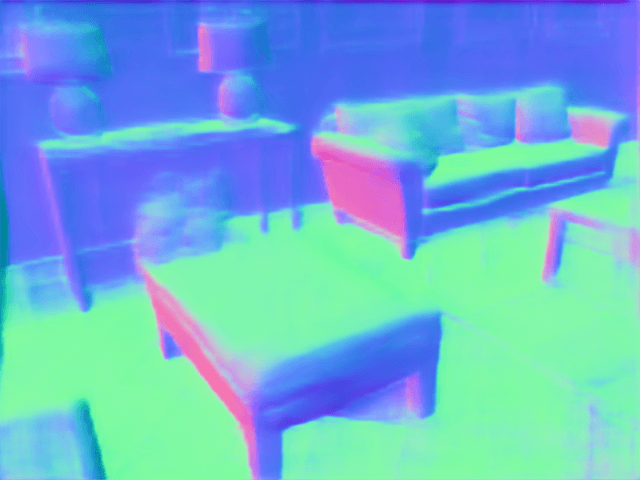} &
			\includegraphics[width=1.2\linewidth]{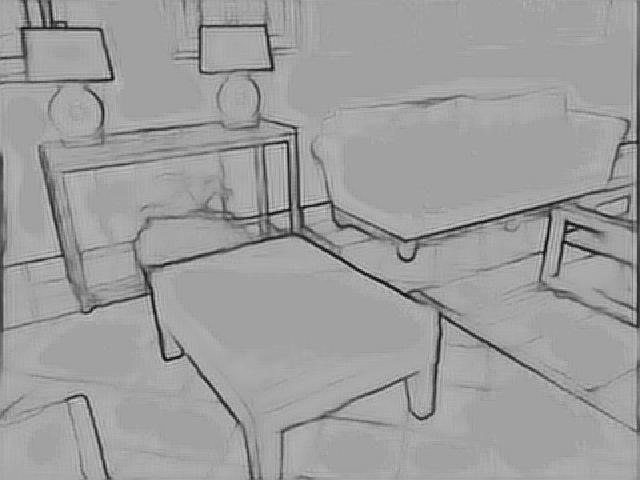}\\

			MP \cite{Matterport3D} &
			\includegraphics[width=1.2\linewidth]{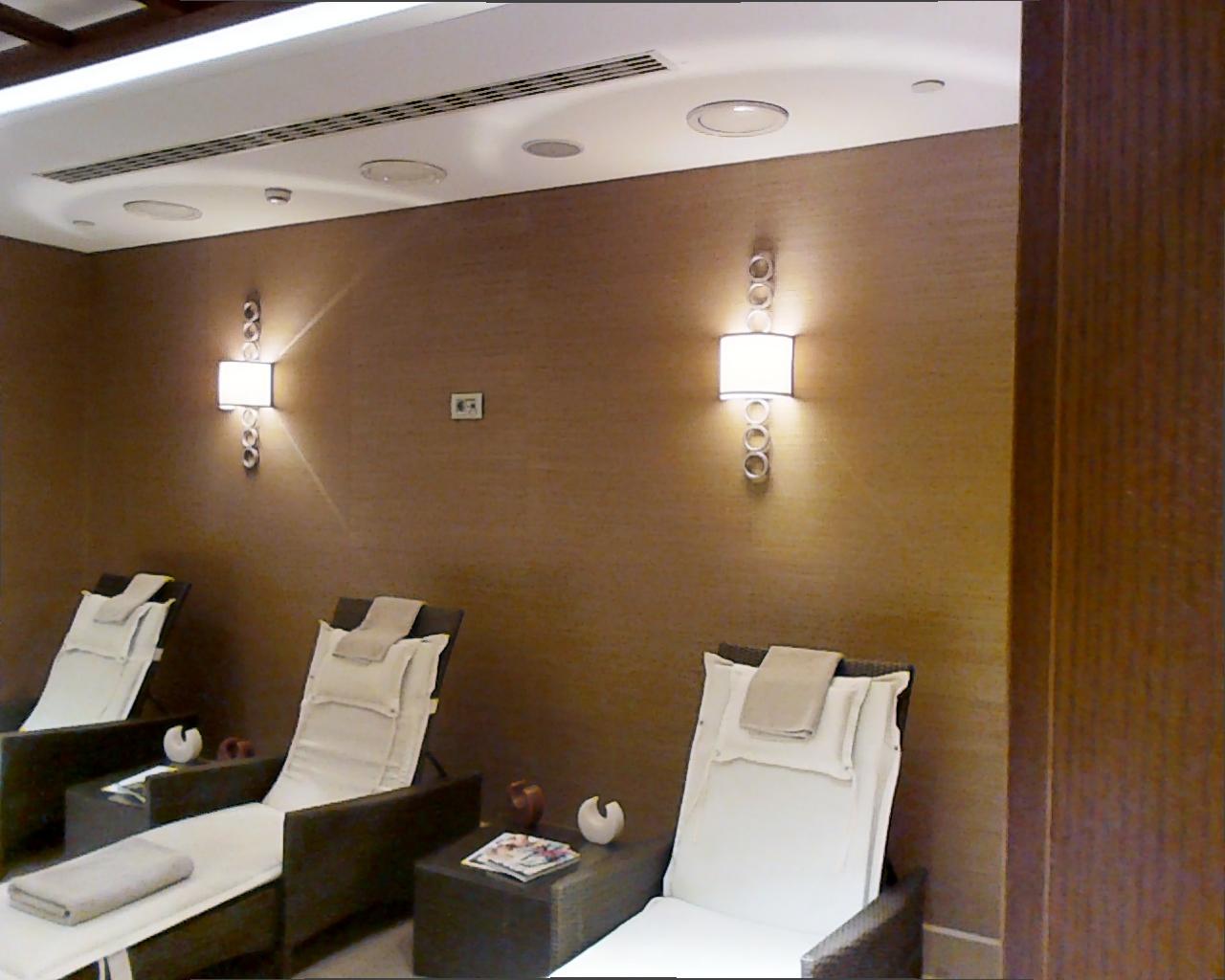} &
			\includegraphics[width=1.2\linewidth]{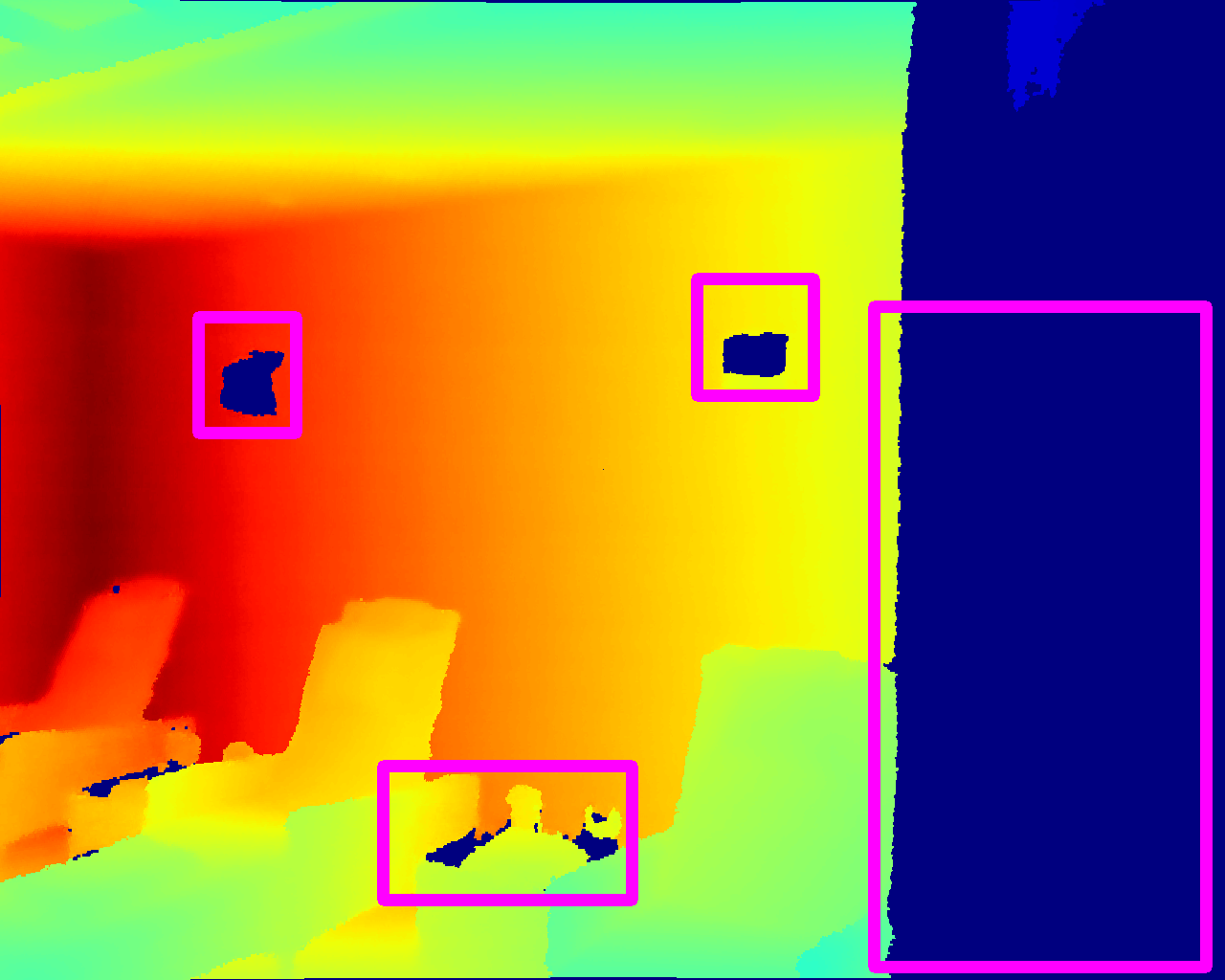}
			 &
			\includegraphics[width=1.2\linewidth]{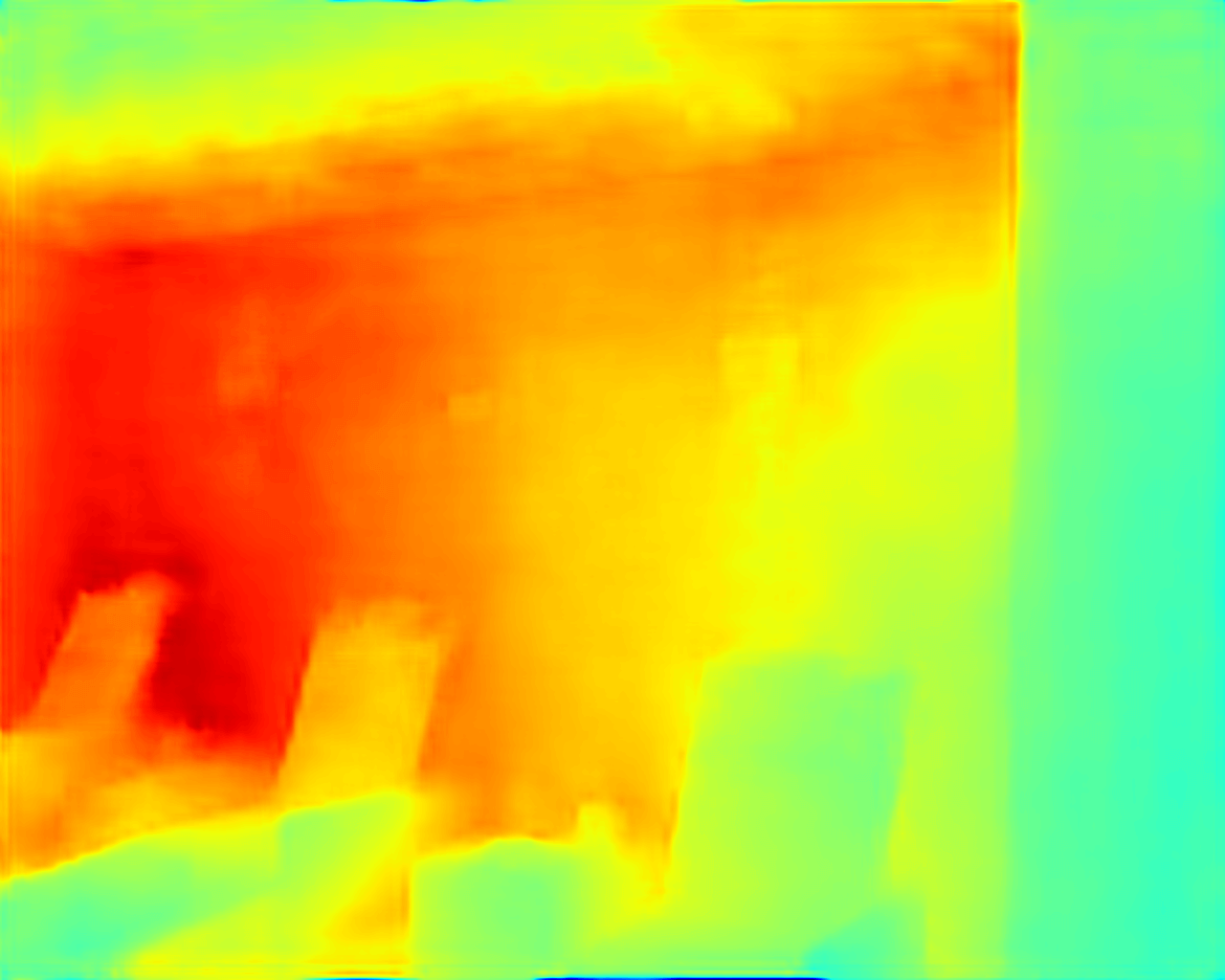} &
			\includegraphics[width=1.2\linewidth]{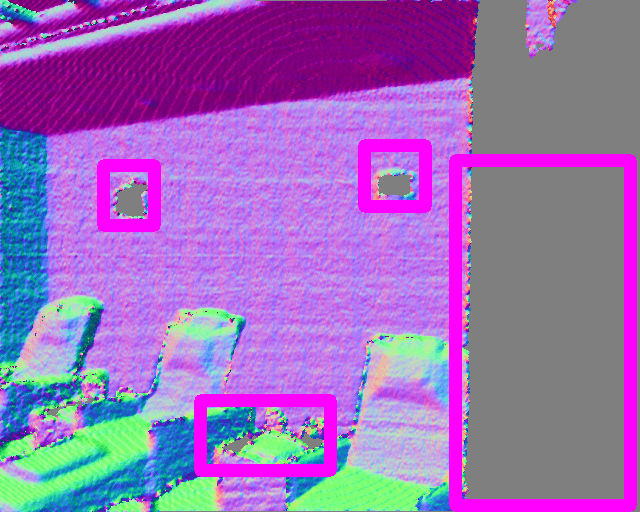}
			 &
			\includegraphics[width=1.2\linewidth]{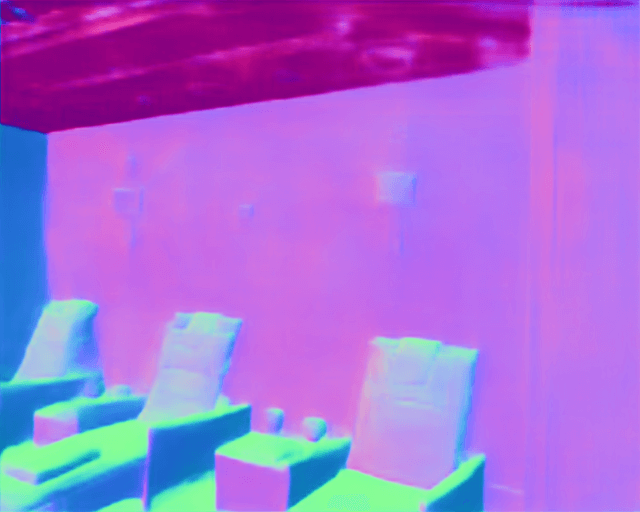}
			 &
			\includegraphics[width=1.2\linewidth]{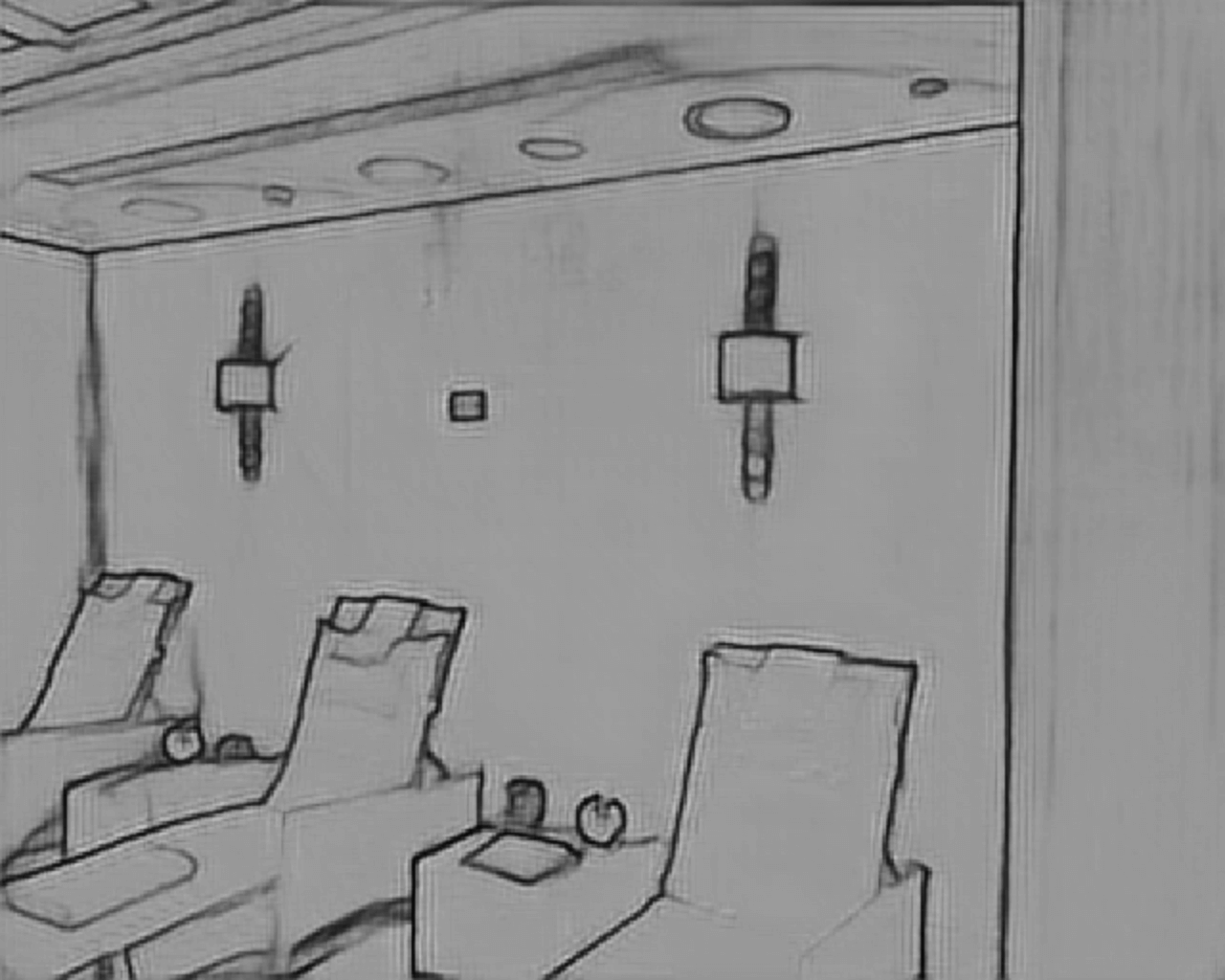}\\

			SUN-RGBD \cite{Sun-rgbd} &
			\includegraphics[width=1.2\linewidth]{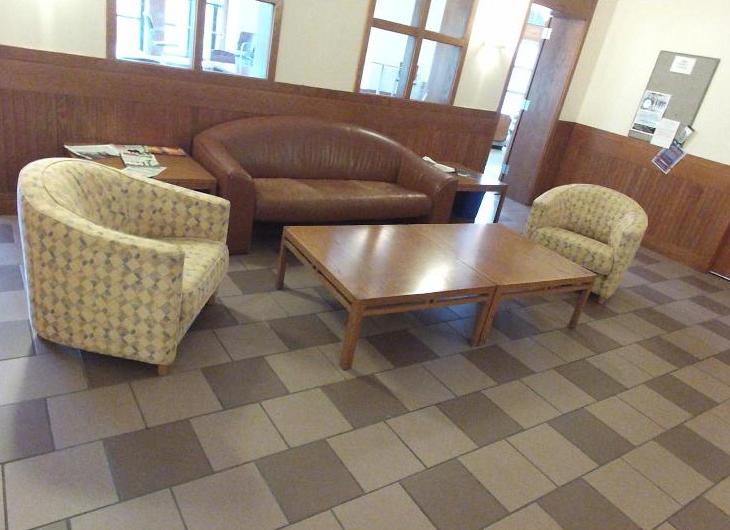} &
			\includegraphics[width=1.2\linewidth]{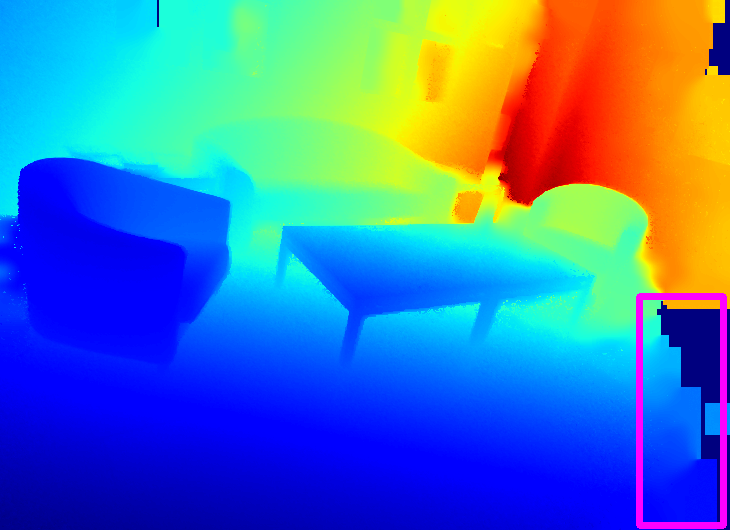}
			 &
			\includegraphics[width=1.2\linewidth]{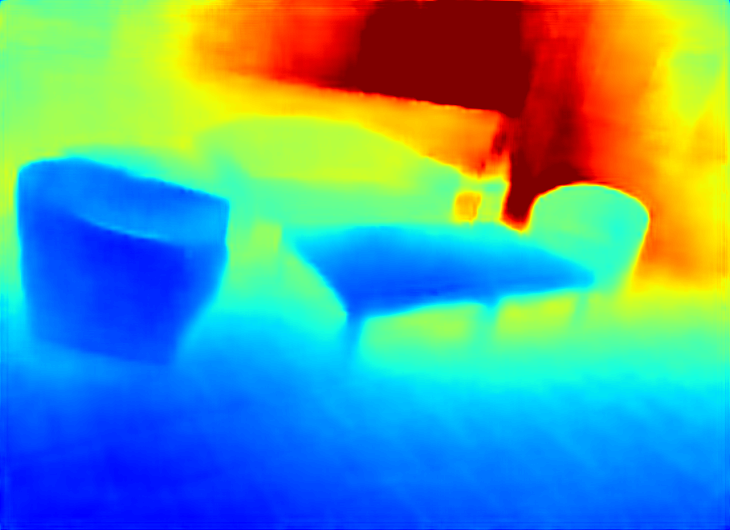} &
			\includegraphics[width=1.2\linewidth]{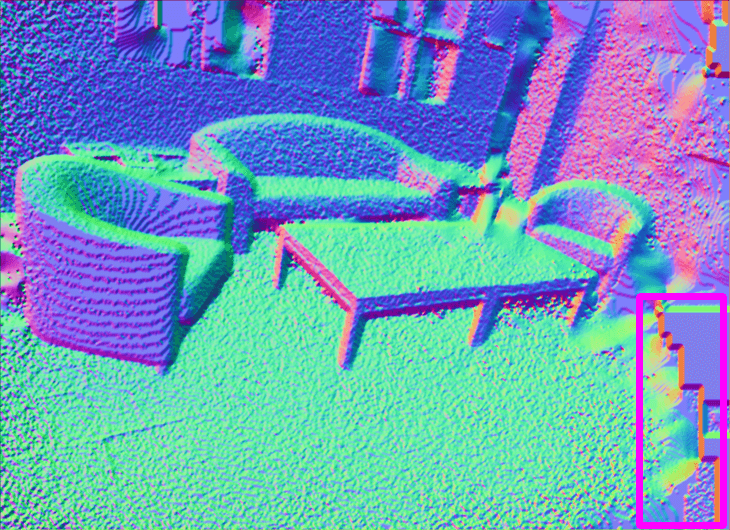}
			 &
			\includegraphics[width=1.2\linewidth]{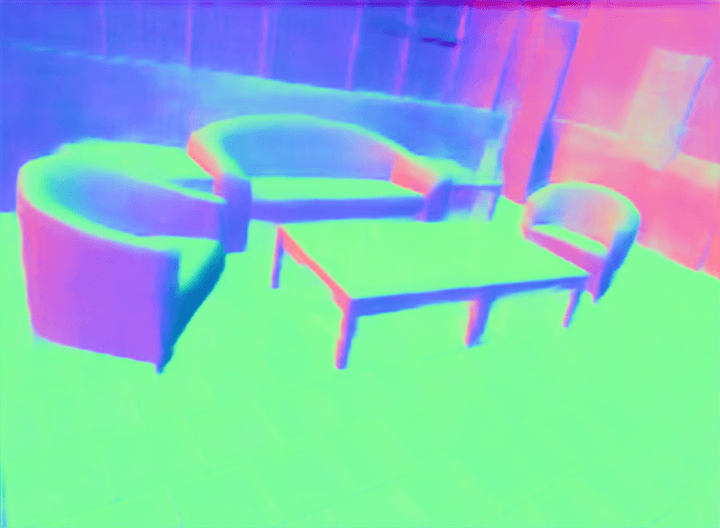} 
			&
			\includegraphics[width=1.2\linewidth]{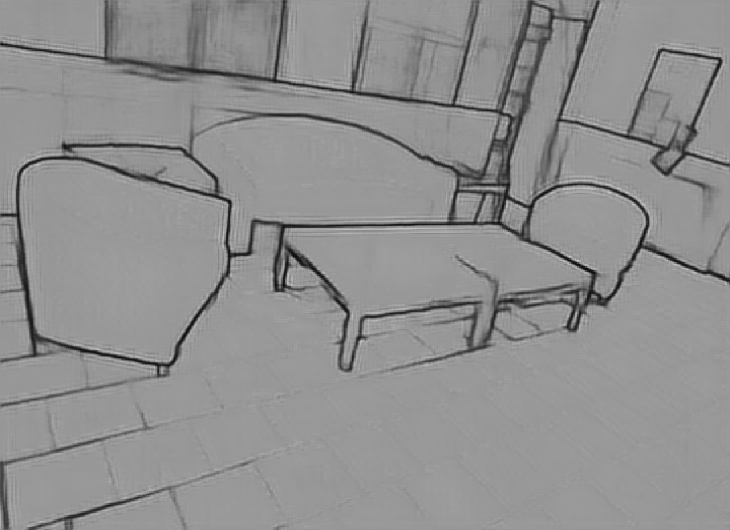}\\
		\end{tabular}
		~\\
		\caption{
			Several predictions on single RGB images from multiple real 
			RGB-D datasets. 
			``MP'' stands for Matterport3D, ``GT'' stands for ground 
			truth and ``Pred'' for prediction.
			We highlight areas where we 
			successfully reconstructed geometry while Kinect depth maps 
			were inaccurate (the chair should be closer than the lamp in first 
			image). 
			Ground truth normals are computed 	
			using code from~\cite{Nyuv2} for NYUv2 
			and~\cite{McCormacSceneNetRGBD} for SUN-RGBD. Normal maps are 
			already provided in Matterport3D.}
		\label{fig:qualitative_predictions}
	\end{center}
\end{figure*}

As shown in  Fig.~\ref{fig:architecture}, we train a  network $f(\bI;\Theta)$ to
predict,  for a  training  color image  $\bI$,  a  depth map  $\bhD$, a map of
occluding contours probabilities $\bhC$, and  a map  $\bhN$ of surface normals. 
Although we focus on high quality depth-maps prediction, our occluding 
contours and normals map can also be used for other applications. Our approach 
generalizes well to various indoor datasets in terms of geometry estimation as 
can be seen in Fig.~\ref{fig:qualitative_predictions}.

\subsection{Training Overview}

We first  train $f$  on the  synthetic dataset  PBRS~\cite{Zhang2016pbrs}, which
provides the ground truth for the depth map  $\bD$, the normals map $\bN$, 
and the binary map of object instance contours $\bC$ for  each training  image 
$\bI$. Since \textit{occluding} contours are not directly provided in the PBRS 
dataset, we choose to use the \textit{object instance} contours $\bC$ as a 
proxy. We argue that on a macroscopic scale, a large proportion of occluding 
contours in an image are due to objects occluding one another.
However, we show that we can also enable our network to learn internal 
occluding contours within objects even without ``pure'' occluding contours 
supervision. Indeed, we make use of constraints on depth map and occluding 
contour predictions $\hat{\bD}$ and $\hat{\bC}$ respectively (see 
Section.~\ref{ssec:consensus_terms} for more details) to enforce the contour 
estimation task to also predict intra-object occluding boundaries.

\begin{figure}[h]
	\begin{center}
			\begin{tabular}{|@{\hskip 0.2in}c@{\hskip 
			0.2in}||cc|}				
				\hline
				RGB & $\sigma^{-}=0.15,$ & 
				$\sigma^{+}=0.3$ \\ 
				\includegraphics[width=0.36\linewidth]{images/nyuv2_ob/88_img.png}
				 & 
				\multicolumn{2}{c|}{\includegraphics[width=0.36\linewidth]{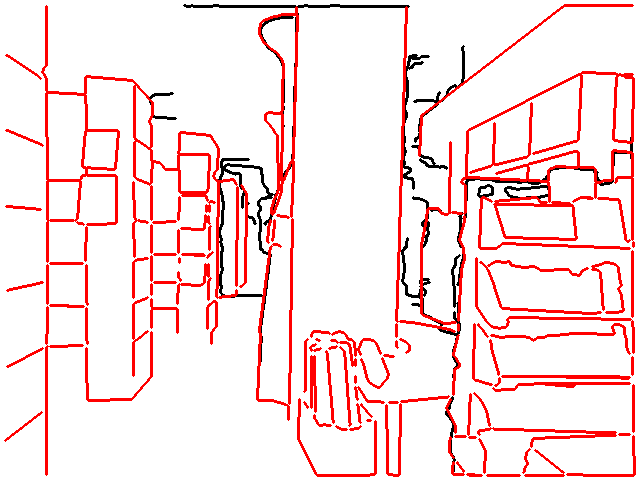}}
				\\								
				\hline
				Depth & $\sigma^{-}=0.01,$ & $\sigma^{+}=0.1$ \\				
				\includegraphics[width=0.36\linewidth]{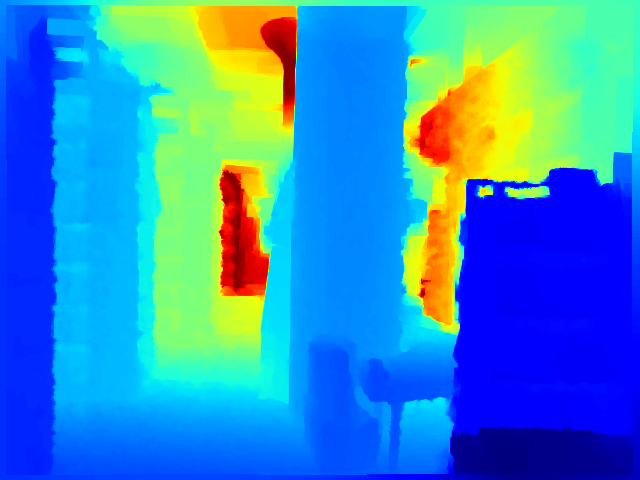}
				 &
				\multicolumn{2}{c|}{\includegraphics[width=0.36\linewidth]{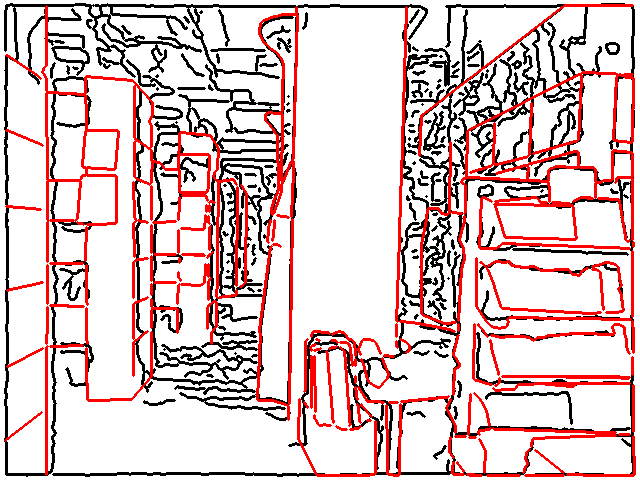}}
				 \\
				\hline
		\end{tabular}
	\end{center}
	\caption{A RGB-D sample of NYUv2-Depth for which we manually annotated 
	occluding contours in NYUv2-OC, (in red lines). We show in black the edges 
	detected on ground truth Kinect-v1 depth map using different Canny detector 
	parameters ($\sigma^{-}$  and $\sigma^{+}$ denote low and high threshold 
	respectively). Highly permissive detectors often yield many spurious 
	contours, whereas restrictive ones miss many true contours. Automatic 
	occluding contours extraction from Kinect depth maps is therefore 
	unreliable for extraction of ground truth occluding contours, motivating 
	our 	manually annotated NYUv2-OC dataset.}
	\label{fig:multi_canny}	
\end{figure}

We then finetune $f$ on the NYUv2-Depth dataset without direct supervision 
on the  occluding contours or normals  ($\calL_\boundaries$ and $\calL_\normals$
described  below): Even  though  \cite{LadickyCVPR14}  and \cite{Nyuv2}  produce
ground  truth normals  map with  different estimation  methods operating  on the
Kinect-v1  depth  maps, their  output  results  are generally  noisy.  Occluding
contours are not given in the  original NYUv2-Depth dataset.  Although one could
automatically extract  them using edge  detectors~\cite{Canny,StructuredEdge} on
depth maps, such extraction is very  sensitive to the detector's parameters (see
Figure~\ref{fig:multi_canny}).  Instead,  we introduce consensus terms  that
explicitly   constrain  the   predicted  contours,   normals  and   depth  maps
together~($\calL_\dc$ and $\calL_\dn$ described below) at training time.

At test-time, we can choose to use only the depth stream of $f$ if we are not
interested in the normals nor the boundaries, making inference very fast.

\subsection{Loss Function}

We estimate the parameters $\Theta$ of network $f$ by minimizing the following 
loss function over all the training images:
\begin{align}
\calL = &\; \lambda_\depth \calL_\depth(\bD, \bhD) + \lambda_\boundaries  
\calL_\boundaries(\bC, \bhC) + \lambda_\normals \calL_\normals(\bN, \bhN) \; 
+ \nonumber\\
&\; \calL_\dc(\bhD, \bhC) + \calL_\dn(\bhD, \bhN) \>,
\label{eq:loss_function}
\end{align}

where
\begin{itemize}
\item $\calL_\depth$, $\calL_\boundaries$, and $\calL_\normals$ are  
supervision terms for the depth, the occluding contours, and the normals 
respectively. We     adjust    weights     $\lambda_\depth$,   
$\lambda_\boundaries$,     and $\lambda_\normals$ during  training so  that we 
focus  first on  learning local geometry     (normals     and     
boundaries)    then     on     depth.      See 
Section~\ref{ssec:implementation}  for  more details.
\item $\calL_\dc$ and $\calL_\dn$ introduce constraints between the predicted
  depth map and the predicted contours, and between the predicted depth map and the
  predicted normals respectively.
\end{itemize}
We detail these losses below. All losses are computed using only valid pixel 
locations.  The PBRS synthetic dataset provides such a mask.  When finetuning 
on NYUv2-Depth, we mask out the white pixels on the 
images border.


\subsection{Supervision Terms $\calL_\depth$, $\calL_\boundaries$, and 
$\calL_\normals$}

The supervision  terms on  the predicted  depth and normal  maps are  drawn from
previous  works  on monocular  depth  prediction.   For  our term  on  occluding
contours prediction, we rely on previous work for edge prediction.

\paragraph{Depth prediction loss $\calL_\depth$.}

As in recent  works, our loss on depth prediction  applies to log-distances.  We
use  the \BerHu~loss  function~\cite{BerHuOwen,BerHuZwald}, as  it  was
shown in \cite{Laina2016DeeperDP}  to result in faster converging
and better solutions:
\begin{equation}
  \begin{aligned}
    \calL_\depth(\bD, \bhD) &= \frac{1}{N}\sum_i \BerHu(\log(\hDi) - 
    \log(D_i))\\&+ \frac{1}{N} \sum_i \|\nabla\log(\hDi) -
    \nabla\log(D_i))\|^2\> .
  \end{aligned}
\end{equation}
The sum is  over all the $N$  valid pixel locations.  The  \BerHu~(also known as
reverse Huber) function is  defined as a $L_2$ loss for  large deviations, and a
$L_1$  loss for  small ones.   As in~\cite{Laina2016DeeperDP},  we take  the $c$
parameter  of   the  \BerHu~function  as  $c=\frac{1}{5}   \max_i(|\log(\hDi)  -
\log(D_i)|)$.

\paragraph{Occluding contours prediction loss $\calL_\boundaries$.}

We use the recent attention loss from \cite{DOOBNet}, which was developed for 2d
edge detection, to learn to predict the occluding contours. This attention loss
helps dealing with  the imbalance of edge pixels compared to non-edge pixels:
\begin{equation}
  \AL(\hat{p}, p) = \begin{cases}
    -\alpha\beta^{(1-\hat{p})^\gamma}\log(\hat{p}) &\text{if } p = 1\\
    -(1 - \alpha)\beta^{\hat{p}^\gamma}\log(1-\hat{p}) &\text{else}
  \end{cases}
\end{equation}
where $(\beta, \gamma)$ are hyper-parameters which  we set to the authors values
$(4,  0.5)$, and  $\alpha$  is computed  image  per image  as  the proportion  
of contour pixels. We use this pixel-wise  attention loss to define the 
occluding contours prediction loss:
\begin{equation}
  \calL_\boundaries(\bC, \bhC) = \frac{1}{N} \sum_i \AL(\hCi, C_i) \> .
\end{equation}
As mentioned  above, this loss  is disabled  when finetuning on  the NYUv2-Depth
dataset.

\paragraph{Normals prediction loss $\calL_\normals$.}

For normals prediction, we use a common method introduced by Eigen \emph{et 
al.}~\cite{Eigen2015PredictingDS} which is to minimize, for all valid pixels 
$i$, the angle between the predicted normals  $\bhN_i$ and their ground 
truth counterpart $\bN_i$. This angle minimization is performed by maximizing 
their dot-product. We therefore used the following loss:
\begin{equation}
  \calL_\normals(\bN,\bhN) = \frac{1}{N} \sum_i \left(1 - \dfrac{<\bhN_i,
    \bN_i>}{\|\bhN_i\|\|\bN_i\|}\right) \> .
\end{equation}

This loss  slightly differs from  the one of \cite{Eigen2015PredictingDS}  as we
limit it  to positive values. As  mentioned earlier, this loss  is disabled when
finetuning on the NYUv2-Depth dataset.

\subsection{Consensus Terms $\calL_\dc$ and 
$\calL_\dn$}\label{ssec:consensus_terms}

\paragraph{Depth-contours consensus term.}
In order to force the network to predict  sharp depth edges at  occluding 
contours where strong depth  discontinuities occur, we  propose the following 
loss  between the predicted occluding contours probability map  $\bhC$ and the 
predicted depth map $\bhD$:
\begin{equation}
\begin{aligned}
  &\calL_\dc(\bhD, \bhC) = -\frac{1}{N} \sum_i 
  \log(\hCi)\cdot\|\nabla(\hDi)\|^2\|\Delta(\hDi)\| \\
  &+\mu \left( \|\bhC\| - \frac{1}{N}\sum_i  \log(1-\hCi)\cdot   
  e^{-\|\Delta(\hDi)\|}\right).
\end{aligned}
 \label{eq:depth_edge_consensus}
\end{equation}
This encourages the network to associate  pixels with large depth gradients with
occluding contours:  High-gradient areas will  lead to  a large loss  unless the
occluding contour probability is close to one.
\cite{monodepth17,Heise2013PMHuber}   also   used   this  type   of   edge-aware
gradient-loss, although  they used  it to  impose consensus  between photometric
gradients and depth gradients. However, relying on 
photometric gradients can be dangerous: textured areas can exhibit strong image 
gradients without strong depth gradients.

\paragraph{Depth-normals consensus loss.}
\label{par:surface_consensus}

Depth and normals are two highly correlated entities. Thus, to impose 
geometric  consistency during prediction between the  normal and depth
predictions $\bhD$ and $\bhN$, we use the following loss:

\begin{equation}
\calL_\dn(\bhD, \bhN) = \frac{1}{N} \sum_i\left(1 -
\dfrac{<\hat{\bu_i}, \hat{\bn_i}>}{\|\hat{\bu_i}\|\|\hat{\bn_i}\|}\right) \> ,
\label{eq:normals_depth_consensus}
\end{equation}

where $\hat{\bn_i} =(\hat{n_x^i}, \hat{n_y^i})^T$ is extracted from the
3D vector $\bhN_i=(\hat{n_x^i}, \hat{n_y^i}, \hat{n_z^i})^T$, and 
$\hat{\bu_i}=(\partial_x \hDi, \partial_y \hDi)$ is computed as the 2D 
gradient of the depth map estimate using finite differences. This term enforces 
consistency between the normals and depth predictions in a similar fashion as 
in~\cite{WangSurgeNIPS16, Yang2018lego, Fei2018GeoSupervisedVD}.
However, our formulation of depth-normals consensus is much simpler than those 
proposed in previous works as they express their constraint in 3D world 
coordinates, thus requiring the camera calibration matrix. Instead, we only 
assume that orthographic projection holds, which is a good first order 
assumption~\cite{Wu_LineIntegration88}.

Imposing this constraint during finetuning allows us to
constrain normals, and depth, even when the ground truth
normals $\bN$ are not available (or accurate enough for our application).

\section{Experiments}

We evaluate our  method and compare it to previous  work using standard metrics,
as well as the depth boundary  edge~(DBE) accuracy  metric introduced by  Koch 
et al.~\cite{Koch2018EvaluationOC}  (see  following  
Section~\ref{ssec:evaluation} and Eq.~\eqref{eq:dbe_accuracy} for more 
details).  We show that our method
achieves the best trade-off  between global reconstruction error and DBE.

\subsection{Implementation Details}\label{ssec:implementation}

We implement our work in Pytorch and make our pretrained weight, training and 
evaluation code publicly 
available.\footnote{~\url{www.github.com/MichaelRamamonjisoa/SharpNet}} 
Both training and evaluation are done on a single high-end NVIDIA GTX 1080 Ti 
GPU.

\paragraph{Datasets.}
We first train our network on the synthetic PBRS~\cite{Zhang2016pbrs} dataset, 
using depth and normals maps annotations, along with object instance boundaries 
maps which we use as a proxy to occluding contours annotations. We split the 
PBRS dataset in \mbox{training/validation/test} sets using a 80\%/10\%/10\% 
ratio. We then finetune our network on the NYUv2-Depth training set using only 
depth data. Finally, we use the NYUv2-Depth validation set for depth evaluation 
and our new NYUv2-OC for occluding contours accuracy evaluation.

\paragraph{Training.} 

Training a multi-task network requires some caution: Since several loss terms
are involved, and in particular one for each task, one should pay special
attention to any suboptimal solution for one task due to `over-learning' 
another. To monitor each task individually, we monitor each individual loss 
along with the global training loss and make sure that all of them decrease 
during training. When setting all loss coefficients equal to one, we noticed 
that the normals loss $\calL_{normals}$ decreased faster than others. 
Similarly, we found that learning boundaries was much faster than learning 
depth. As \cite{zhang2018deepdepth}, we also argue that this is because local 
features such as contours or local planes, \emph{i.e.} where normals are 
constant, are easier to learn since they appear in almost all training 
examples. Training depth, however, requires the network to exploit context data 
such as room layout in order to regress a globally consistent depth map.

Based on those observations, we chose to learn the easier tasks first, then
use them as guidance to the more complex task of depth estimation through our
novel consensus loss terms of 
Eqs.~\eqref{eq:normals_depth_consensus}~and~\eqref{eq:depth_edge_consensus}. 
For finetuning on real data with the NYUv2 dataset, we first disabled the 
consensus terms and froze the contours and normals decoders in order to first 
bridge the depth domain gap between PBRS and NYUv2. After convergence, we 
finetuned the network again with consensus terms back on, which helped 
enhancing predictions by ensuring consistency between geometric entities. 
We found that it was necessary to freeze the normals and contours decoders 
during finetuning to prevent their predictions $\bhC$ and $\bhN$ from degrading 
until being unable to play their geometry guidance role.
We argue that this is due to (1) a larger synthetic-to-real domain gap for depth than for contours and normals, and (2) noisy depth ground truth with some inaccuracies along occluding contours and crease along walls.
We therefore relied on the ResNet50 encoder to learn a representation which 
produces geometrically consistent predictions $\bhC$, $\bhN$ and $\bhD$.

\subsection{Evaluation Method}\label{ssec:evaluation}

We evaluate our method on the benchmark dataset NYUv2 Depth~\cite{Nyuv2}. The most common metrics are: Thresholded accuracies $(\delta_1,\delta_2, 
\delta_3)$, linear and logarithmic Root Mean Squared Error RMSE$_{lin}$ and 
RMSE$_{log}$, Absolute Relative difference $rel$, and logarithmic error 
$\log_{10}$.

\begin{table*}[h]
	\footnotesize
	\begin{center}
			\begin{tabular}{|@{\hspace{0.15em}}c@{\hspace{0.15em}}
					@{\hspace{0.15em}}||c@{\hspace{0.15em}}
					@{\hspace{0.15em}}c@{\hspace{0.15em}}
					@{\hspace{0.15em}}c|@{\hspace{0.15em}}
					@{\hspace{0.15em}}c@{\hspace{0.15em}}
					@{\hspace{0.15em}}c@{\hspace{0.15em}}
					@{\hspace{0.15em}}c@{\hspace{0.15em}}
					@{\hspace{0.15em}}c|@{\hspace{0.15em}}
					@{\hspace{0.15em}}c@{\hspace{0.15em}}
					@{\hspace{0.15em}}c@{\hspace{0.15em}}
					@{\hspace{0.15em}}c@{\hspace{0.15em}}
					@{\hspace{0.15em}}c@{\hspace{0.11em}}|
					}
				\hline
				~ & 
				\multicolumn{7}{c|@{\hspace{0.3em}}}{Evaluated
				 on full 
				NYUv2-Depth} 
				& 
				\multicolumn{4}{@{\hspace{0.15em}}c@{\hspace{0.11em}}|}{Evaluated
				 
				on our 
				NYUv2-OC} 
				\\					
				\cline{2-12}
				Method & 
				\multicolumn{3}{c|@{\hspace{0.3em}}}{
				Accuracy 
				$\uparrow 
					(\delta_i = 
					1.25^i)$}  
				& 
				\multicolumn{4}{c|@{\hspace{0.3em}}}{Error
					 
					$\downarrow$} 
				&			
				\multicolumn{4}{@{\hspace{0.15em}}c@{\hspace{0.15em}}|}{$\epsilon_{DBE}^{acc}$
				 
					$\downarrow$(px) $\{\sigma^{-}, \sigma^{+}\}$}\\
				~ & $\delta_1$ & $\delta_2$ & $\delta_3$ & rel & 
				$\log_{10}$ & RMSE (lin) & RMSE (log) & $\{0.1, 0.2\}$ & 
				$\{0.01, 0.1\}$ & $\{0.005, 0.06\}$ & $\{0.03, 0.05\}$ \\
				\hline\hline
				Eigen \textit{et al.}~\cite{Eigen2015PredictingDS} 
				(VGG) & 0.766 & 0.949 & 0.988 & 0.195 & 0.068 & 0.660 & 0.217 & 
				2.895 & 3.065 & \ul{3.199} & \ul{3.203}\\
				Eigen \textit{et al.}~\cite{Eigen2015PredictingDS} 
				(AlexNet) 
				& 0.690$^*$ & 0.911$^*$ & 0.977$^*$ & 0.250$^*$ & 0.082$^*$ & 0.755$^*$ & 0.259$^*$ & 
				\ul{2.840} & \ul{3.029} & 3.202 & 3.242 \\
				Laina \textit{et al.} \cite{Laina2016DeeperDP} 
				& 0.818 & 0.955 & 0.988 & 0.170 & 0.059 & 0.602 & 0.200 & 3.901 & 4.033 & 4.116 & 4.133 \\
				Fu \textit{et al.} \cite{FuCVPR18-DORN} & 
				0.850 & 0.957 & 0.985 & 0.150 & 0.052 & 0.578 & 0.194 
				& 3.714 & 3.754 & 4.040 & 4.062 \\
				Jiao \textit{et al.} \cite{Jiao2018LookDI} & \textbf{0.909} & \textbf{0.981} & \textbf{0.995} & \textbf{0.133} & \textbf{0.042} & \textbf{0.401} & \textbf{0.146} & 
				6.389$^*$ & 4.073$^*$ & 4.179$^*$ & 4.190$^*$ \\
				Ours & \ul{0.888} & \ul{0.979} & \textbf{0.995} & \ul{0.139} & \ul{0.047} & \ul{0.495} & \ul{0.157} &
				\textbf{2.272} &  \textbf{2.629} & \textbf{3.066} & \textbf{3.152} \\
				\hline
		\end{tabular}
	\end{center}
	\caption{Our final evaluation results. Bold and underlined results indicate 
	first and second place respectively. Asterisks indicate the last place. 
	Numerical results might vary from the original papers, as we evaluated 
	all methods with the same code, using only the authors depth map 
	predictions. Results are evaluated in the center crop proposed by 
	\cite{Eigen2015PredictingDS} and clipped depth predictions to range $[0.7m, 
	10m]$.}
	\label{tab:final_evaluation_table}
\end{table*}

\paragraph{NYUv2-Depth benchmark evaluation.}

We have run a comparative study between our method and previous ones, 
summarized in Table~\ref{tab:final_evaluation_table}. 
Since authors evaluating on the NYUv2-Depth benchmark often apply different 
evaluation methods, fair comparison is difficult to perform. For 
instance, \cite{xu2017MS-CRF} and \cite{FuCVPR18-DORN} evaluate on crops 
with regions provided by Eigen \emph{et al.}~\cite{Eigen2015PredictingDS}. 
Some authors also clip resulting depth-maps to the valid depth sensor 
range [0.7m; 10m]. Most importantly, not all the authors make their prediction 
and/or evaluation code publicly available. The authors of \cite{Jiao2018LookDI} 
kindly shared their predictions on the NYUv2-Depth dataset with us, and the 
following evaluation of their method was obtained based on the depth map 
predictions they provided us with. All other mentioned methods have released 
their predictions online.

Fair comparison is ensured by performing evaluation of each method solely using 
its associated depth map predictions and one single evaluation code.

\paragraph{Occluding contours location accuracy.}

To evaluate occluding contours location accuracy, we follow the work 
of Koch \textit{et al.}~\cite{Koch2018EvaluationOC} as they proposed an 
experimental method for such evaluation. Since it is fundamental to examine 
whether predicted depths maps are able to represent all occluding contours as 
depth discontinuities in an accurate way, they analyzed occluding contours 
accuracy performances by detecting edges in predicted and ground 
truth depth maps and comparing those edges.

\begin{figure}[t]		
	\begin{center}
		\includegraphics[width=\linewidth, 
		height=1.5cm]{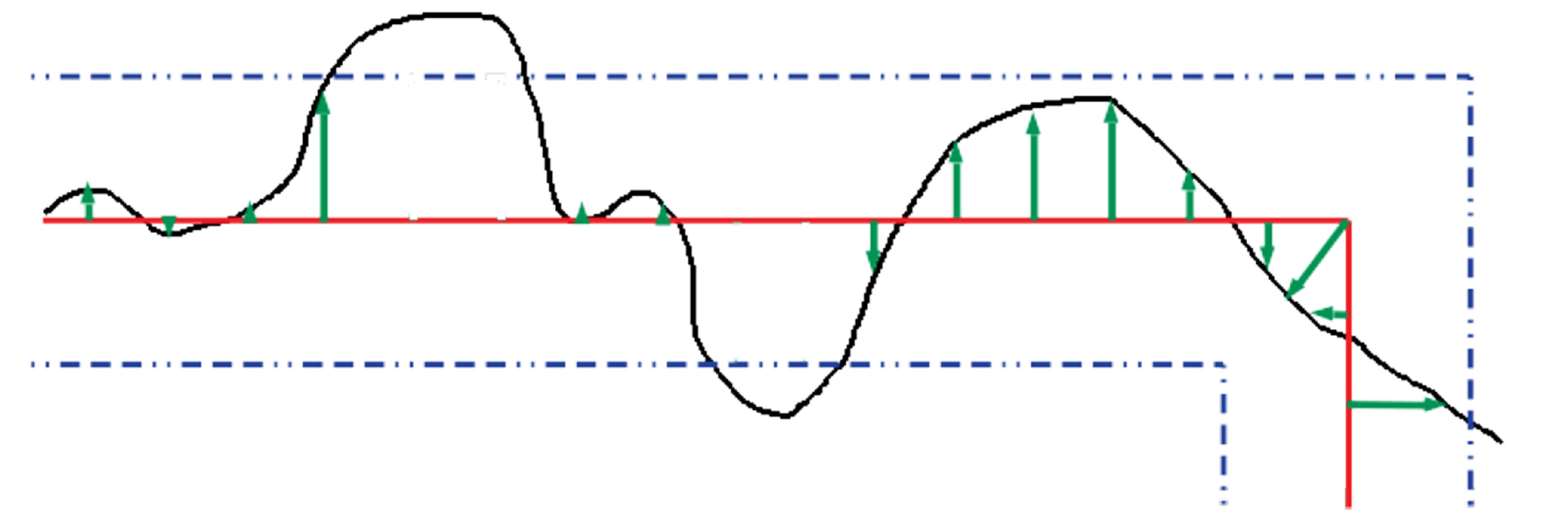}
	\end{center}	
	\caption{The truncated chamfer distance is computed as the sum Euclidean distances $E_i$ (in green) between the detected 
	edge $\widehat{Y}_i$ (in black) and the ground truth edge $Y_i$ (in 
	red). The $E_i$ above 10 pixels (above the blue dashed line) are ignored.}
	\label{fig:dbe_acc_computation}
\end{figure}

\begin{figure*}[t]
	\begin{center}
			\begin{tabular}{@{\hspace{0.2em}}c@{\hspace{0.2em}}
							@{\hspace{0.2em}}c@{\hspace{0.2em}}
							@{\hspace{0.2em}}c@{\hspace{0.2em}}
							@{\hspace{0.2em}}c@{\hspace{0.2em}}
							@{\hspace{0.2em}}c@{\hspace{0.2em}}
							@{\hspace{0.2em}}c@{\hspace{0.2em}}}
				RGB & Laina et al~\cite{Laina2016DeeperDP} & Fu et 
				al~\cite{FuCVPR18-DORN} & Jiao et 
				al~\cite{Jiao2018LookDI} & 
				GT (NYUv2) & SharpNet\\
				\includegraphics[width=0.145\linewidth]{images/nyuv2_ob/1348_img.png}
				 & 
				\includegraphics[width=0.145\linewidth]{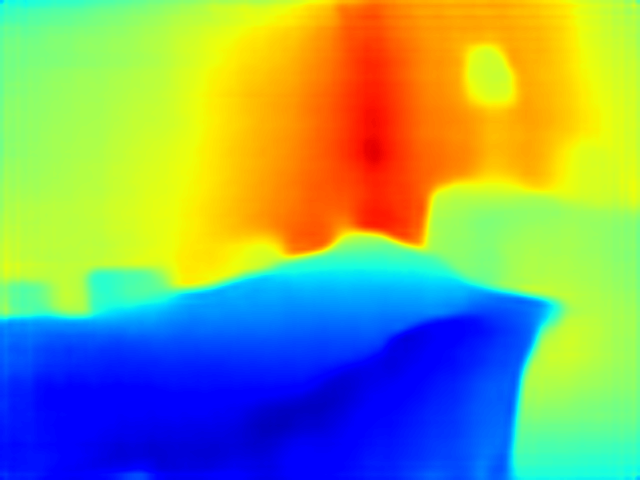}
				 &
				\includegraphics[width=0.145\linewidth]{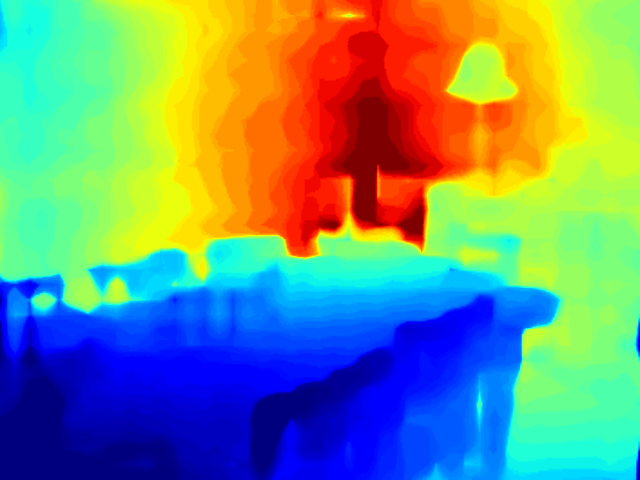}
				 &
				\includegraphics[width=0.145\linewidth]{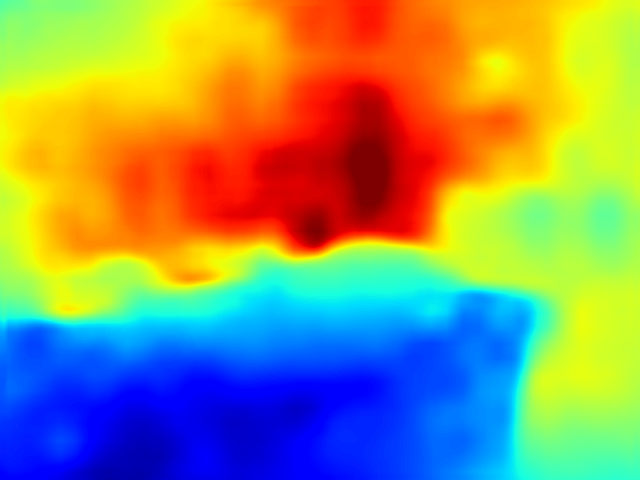}
				 &
				\includegraphics[width=0.145\linewidth]{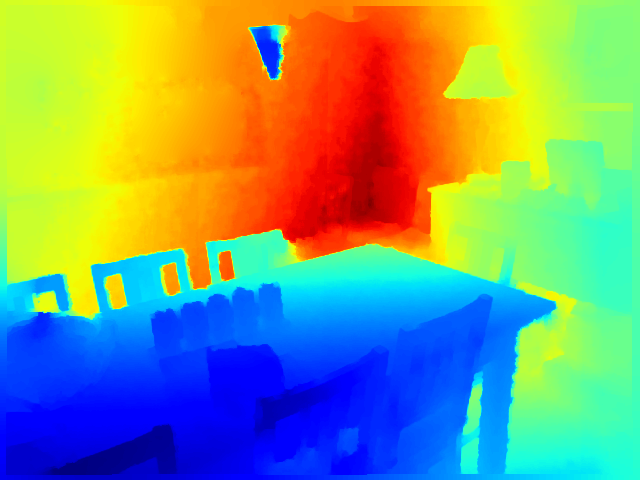}
				 &
				\includegraphics[width=0.145\linewidth]{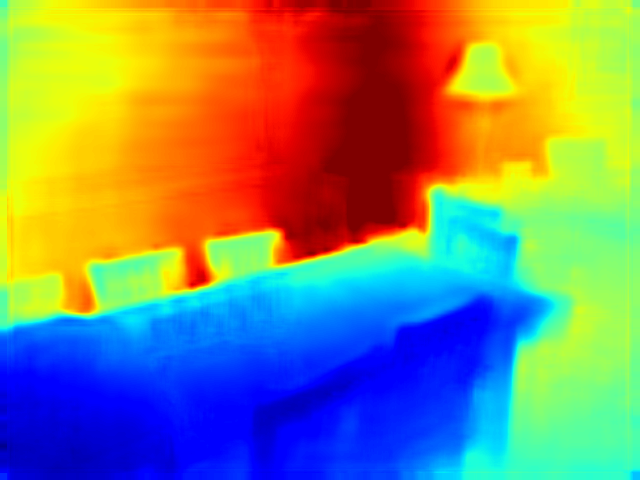}\\
				
				~ & 
				\includegraphics[width=0.145\linewidth]{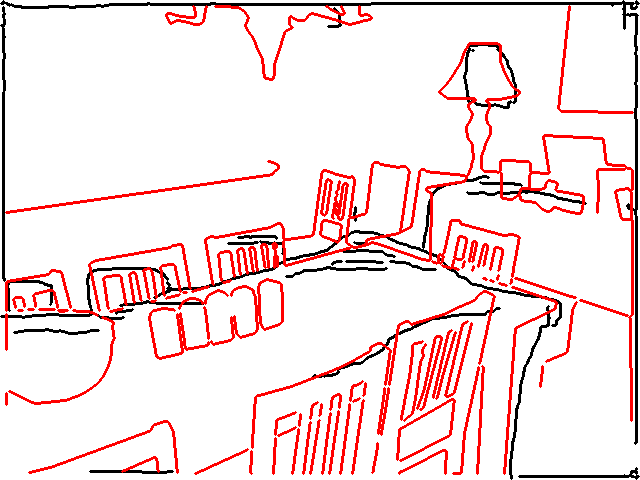}
				 &
				\includegraphics[width=0.145\linewidth]{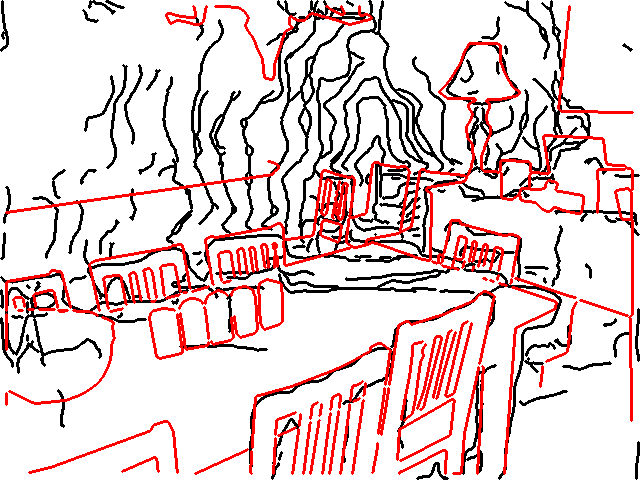}
				 &
				\includegraphics[width=0.145\linewidth]{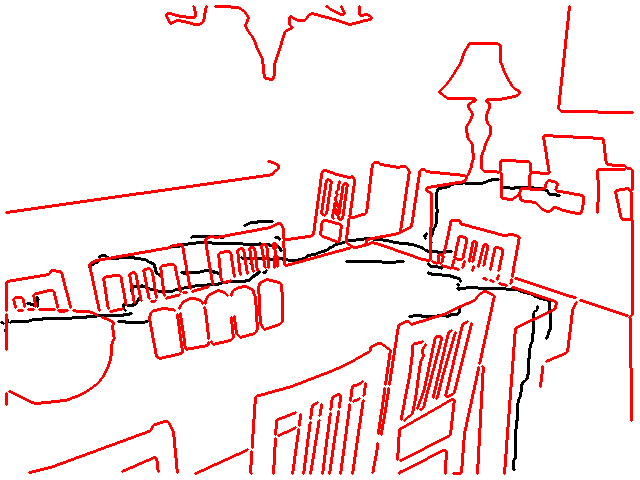}
				 &
				\includegraphics[width=0.145\linewidth]{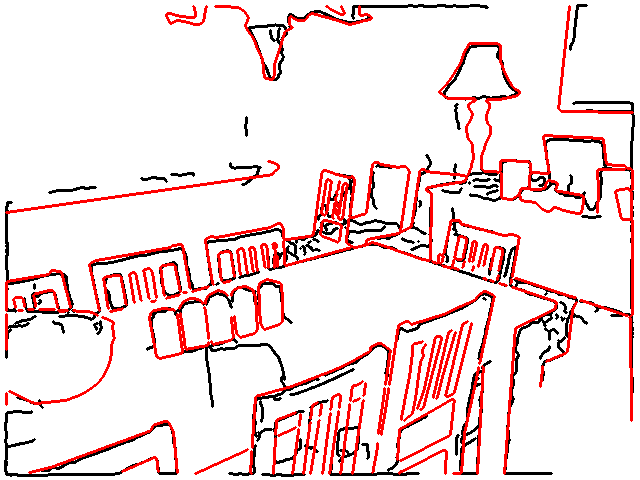}
				 &
				\includegraphics[width=0.145\linewidth]{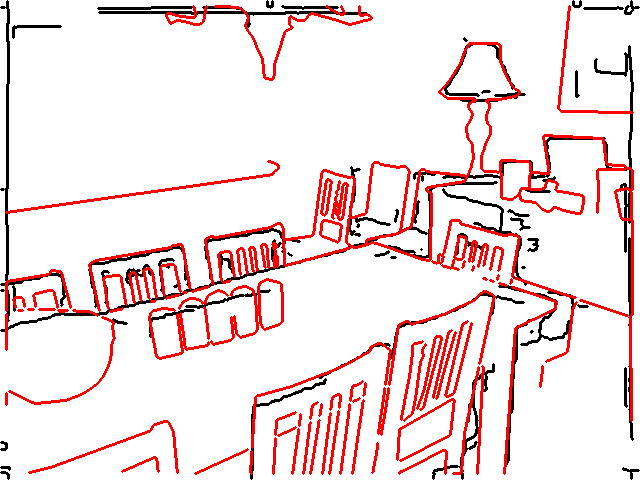}\\

				\includegraphics[width=0.145\linewidth]{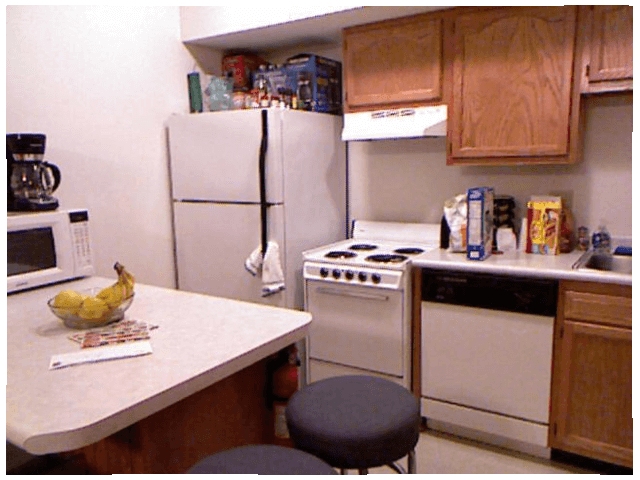}
				 & 
				\includegraphics[width=0.145\linewidth]{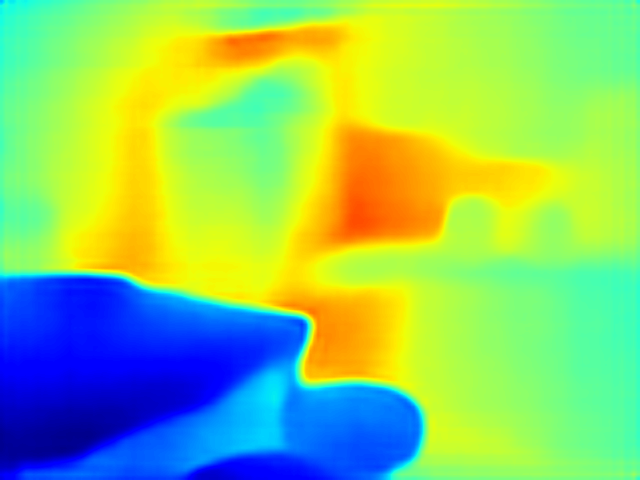}
				 &
				\includegraphics[width=0.145\linewidth]{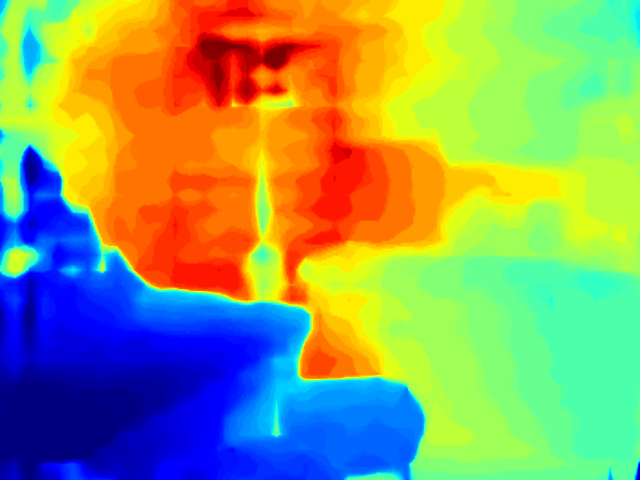}
				 &
				\includegraphics[width=0.145\linewidth]{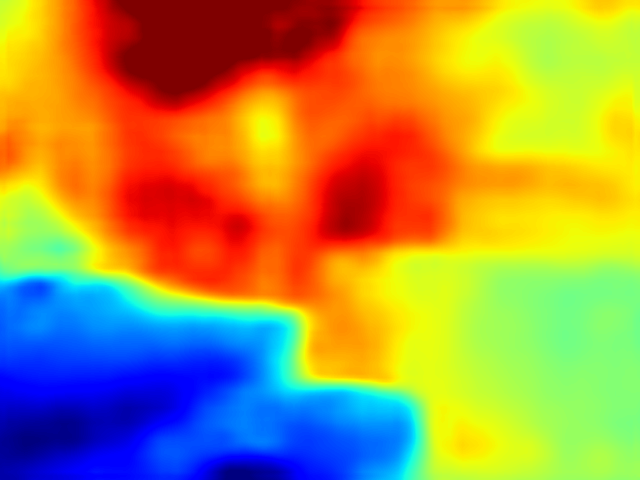}
				 &
				\includegraphics[width=0.145\linewidth]{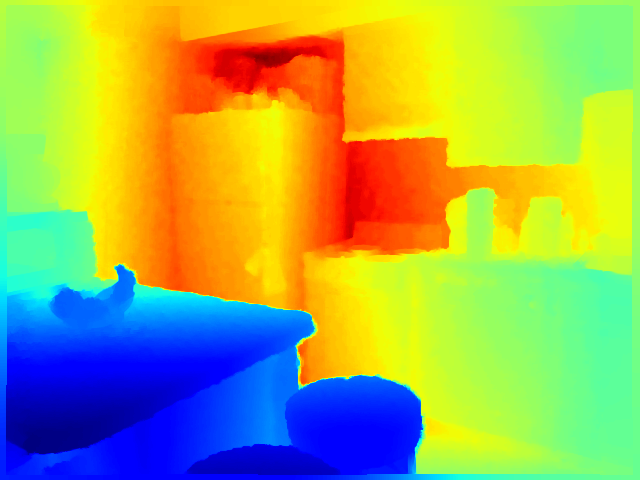}
				 &
				\includegraphics[width=0.145\linewidth]{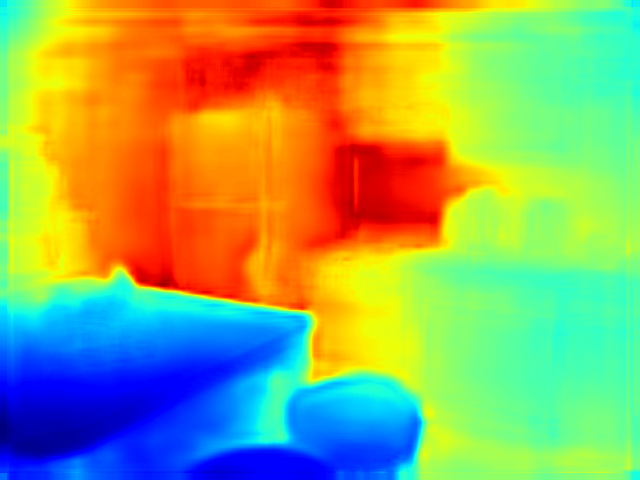}\\
				
				~ & 
				\includegraphics[width=0.145\linewidth]{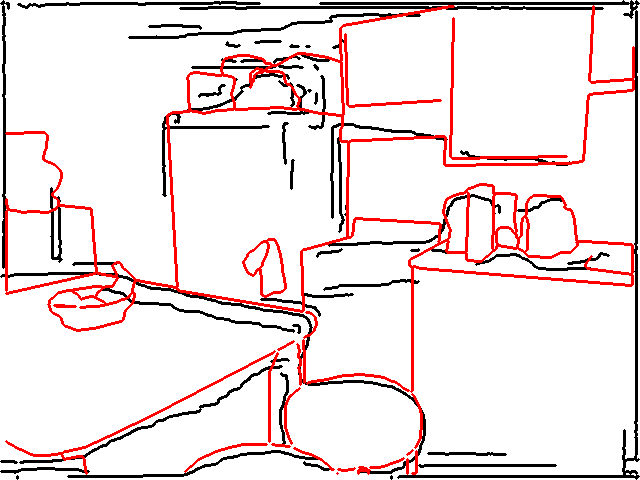}
				 &
				\includegraphics[width=0.145\linewidth]{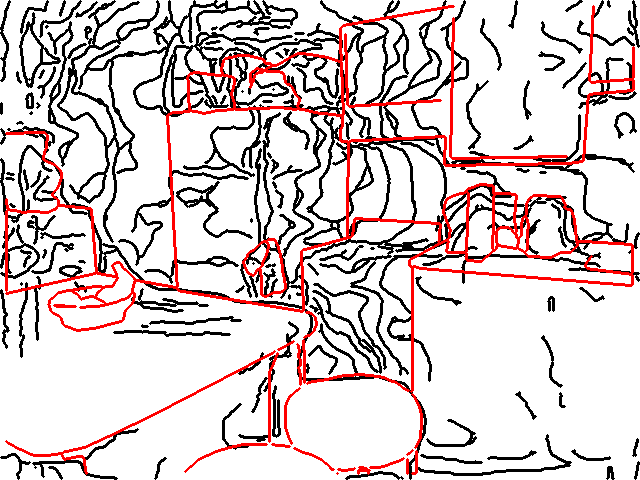}
				 &
				\includegraphics[width=0.145\linewidth]{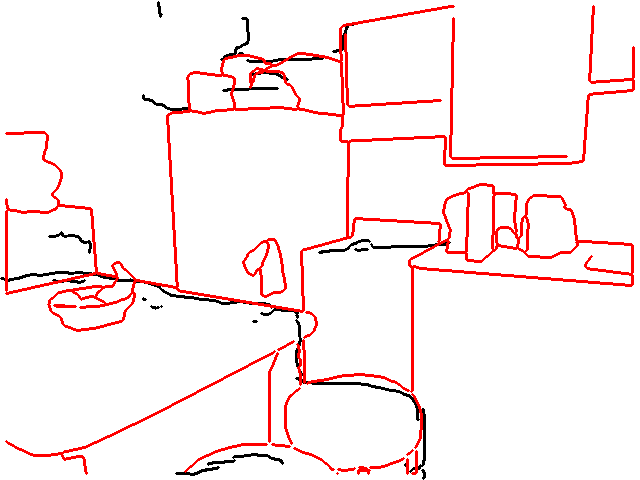}
				 &
				\includegraphics[width=0.145\linewidth]{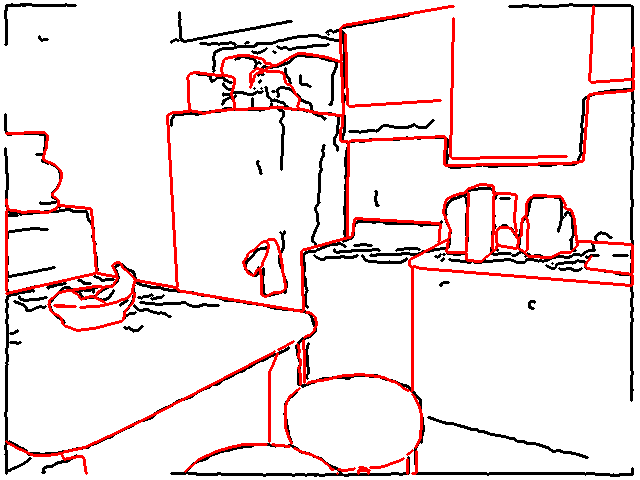}
				 &
				\includegraphics[width=0.145\linewidth]{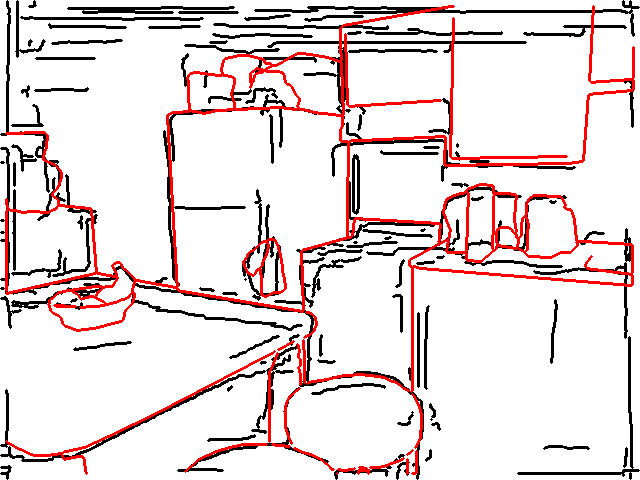}\\

				\includegraphics[width=0.145\linewidth]{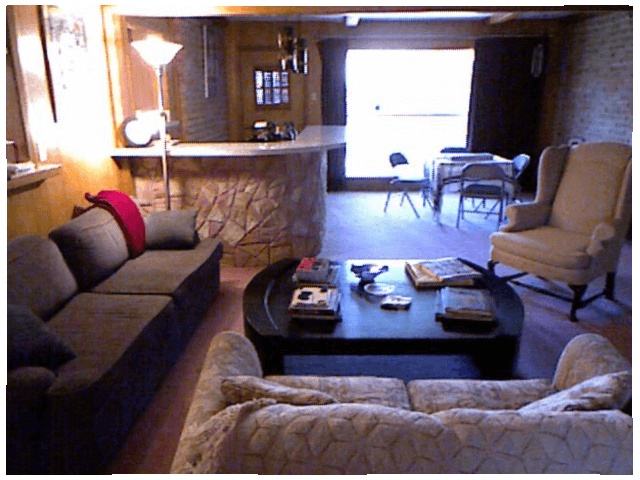}
				 & 
				\includegraphics[width=0.145\linewidth]{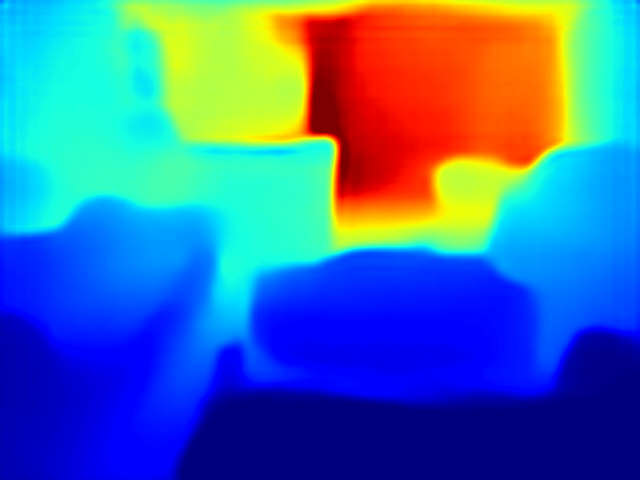}
				 &
				\includegraphics[width=0.145\linewidth]{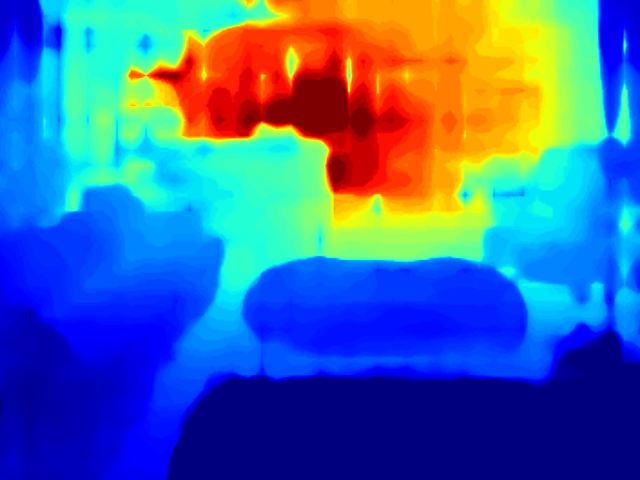}
				 &
				\includegraphics[width=0.145\linewidth]{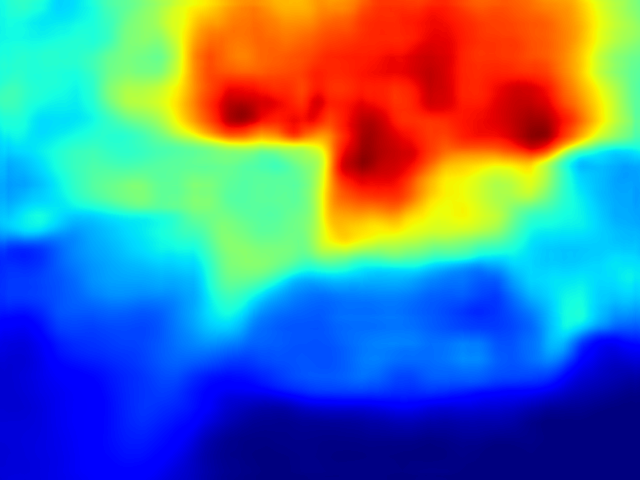}
				 &
				\includegraphics[width=0.145\linewidth]{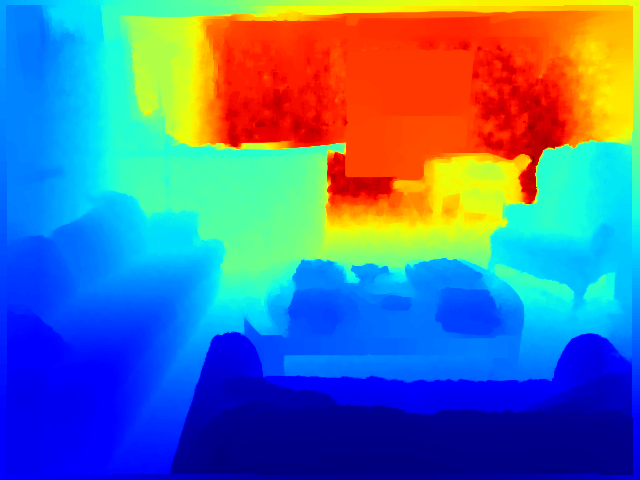}
				 &
				\includegraphics[width=0.145\linewidth]{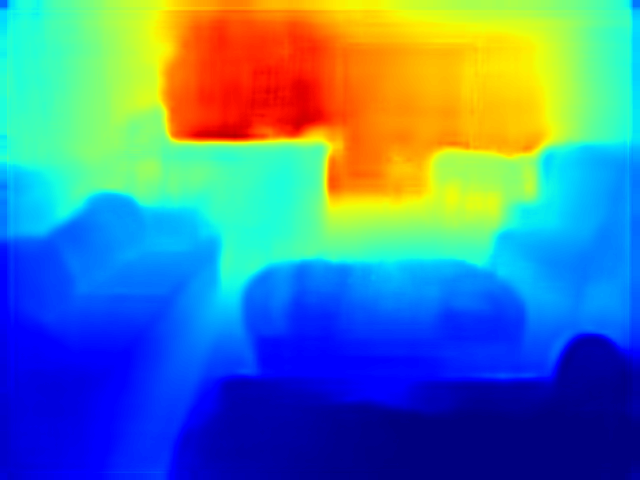}\\
				
				~ & 
				\includegraphics[width=0.145\linewidth]{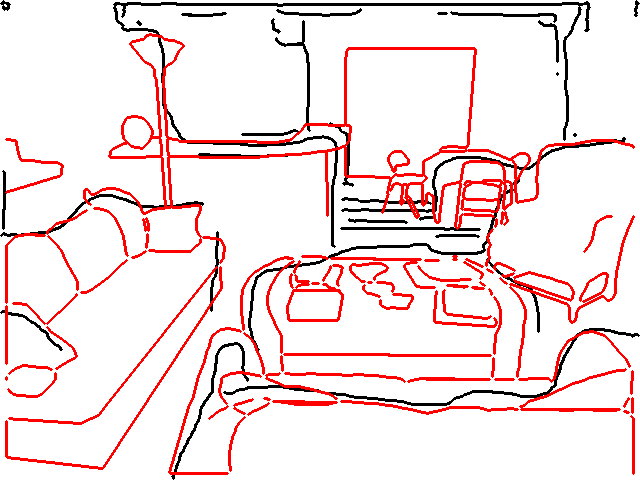}
				 &
				\includegraphics[width=0.145\linewidth]{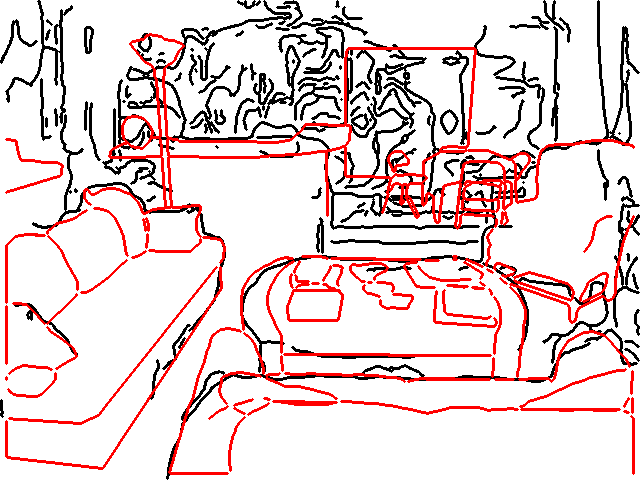}
				 &
				\includegraphics[width=0.145\linewidth]{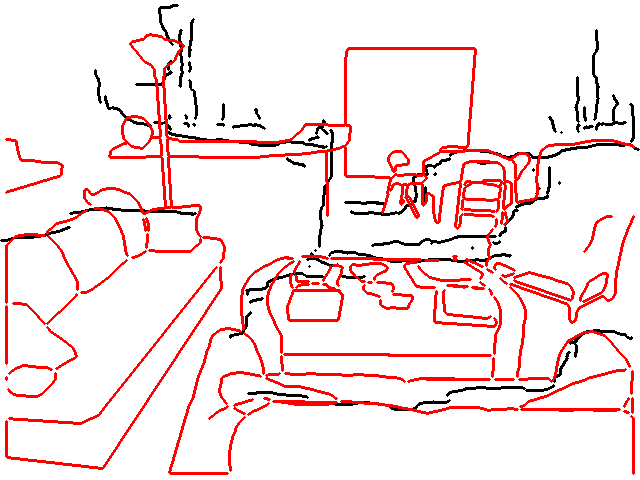}
				 &
				\includegraphics[width=0.145\linewidth]{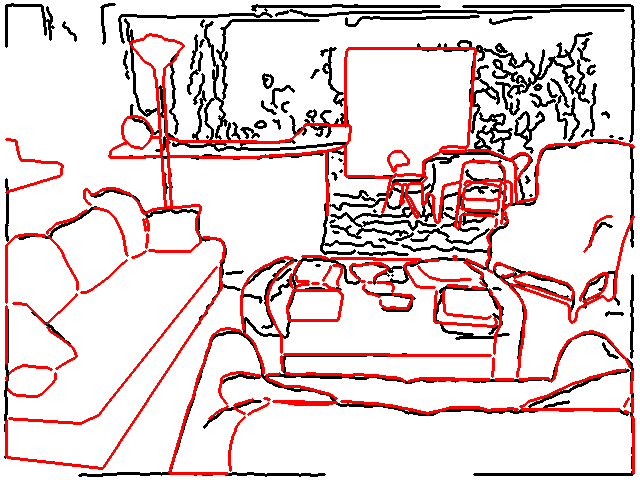}
				 &
				\includegraphics[width=0.145\linewidth]{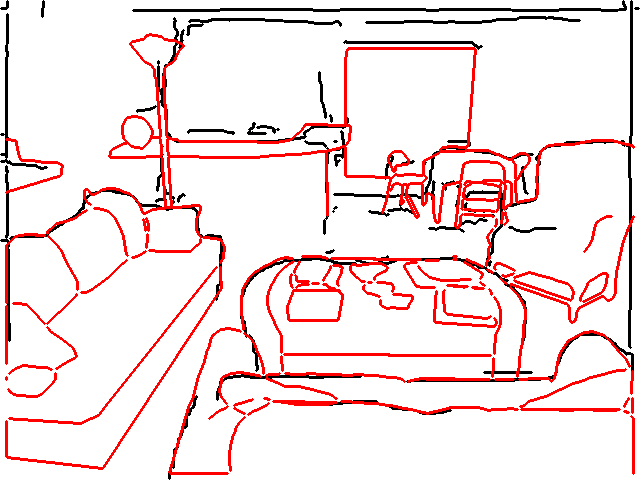}\\
				
		\end{tabular}
	~\\
	\caption{Several examples of images from our NYUv2-OC 
	dataset and their associated depth map estimate for different methods. The 
	second row for each image shows the in black the detected edges on those 
	estimates using a Canny edge detector (in black) with $\sigma^{-}=0.03$ and 
	$\sigma^{+}=0.05$, overlaid on our manually annotated ground truth in red. 
	Our SharpNet method not only creates sharper occluding contours, leading to 
	less spurious and erroneous contours than with \cite{FuCVPR18-DORN} the 
	Kinect-v1 depth-map; it also leads to much better located edges than other 
	methods.}
	\label{fig:DBE}
	\end{center}
\end{figure*}

Since acquired depth maps in the NYUv2-Depth dataset are especially noisy 
around occluding boundaries, we manually annotated a subset of the dataset with 
occluding contours, building our NYUv2-OC dataset, which we used for 
evaluation. Several samples of our NYU-OC dataset are shown in 
Fig.~\ref{fig:multi_canny} and Fig.~\ref{fig:DBE}. In order to evaluate the 
predicted depth maps' $\bD$ quality in terms of occluding contours 
reconstruction, binary edges $\boldsymbol{\widehat{Y}}$ are first extracted 
from $\widehat{\bD}$ with a Canny detector.\footnote{Edges are extracted from 
depth maps with normalized dynamic range.} They are then compared to the ground 
truth annotated binary edges $\boldsymbol{Y}$ from our NYU-OC dataset by 
measuring the a \textit{Truncated Chamfer Distance} (TCD).
Specifically, for each pixel $\widehat{Y}_i$ of $\boldsymbol{\widehat{Y}}$ we 
compute its euclidean distance $E_i$ to the closest edge pixel 
$\widehat{Y}_j=1$. If the distance between $\widehat{Y}_i$ and $\widehat{Y}_j$ 
is bigger than $10$ pixels we set $e_i$ to $0$ in order to evaluate predicted 
edges only around the ground truth edges as seen in 
Fig.~\ref{fig:dbe_acc_computation}. This is done efficiently using 
\textit{Euclidean Distance Transform} on $\boldsymbol{Y}$.
The depth boundary edge (DBE) accuracy is then computed as the mean TCD over 
detected edges $\widehat{Y}_i=1$: 

\begin{equation}
\epsilon^{acc}_{DBE} = \dfrac{1}{\sum\limits_i 
	\widehat{Y}_i}\sum_i E_i \cdot 
\widehat{Y}_i, 
\label{eq:dbe_accuracy}   
\end{equation}

We compare our method against state-of-the-art depth estimation methods using 
this metric and different Canny parameters. Evaluation results are shown in 
Table~\ref{tab:final_evaluation_table}: We outperform all state-of-the-art 
methods on occluding contours accuracy, while being a 
competitive second best on standard depth estimation evaluation 
metrics.

\begin{figure}[t]	
	\begin{center}		
		\includegraphics[width=0.8\linewidth, 
		height=4.4cm]{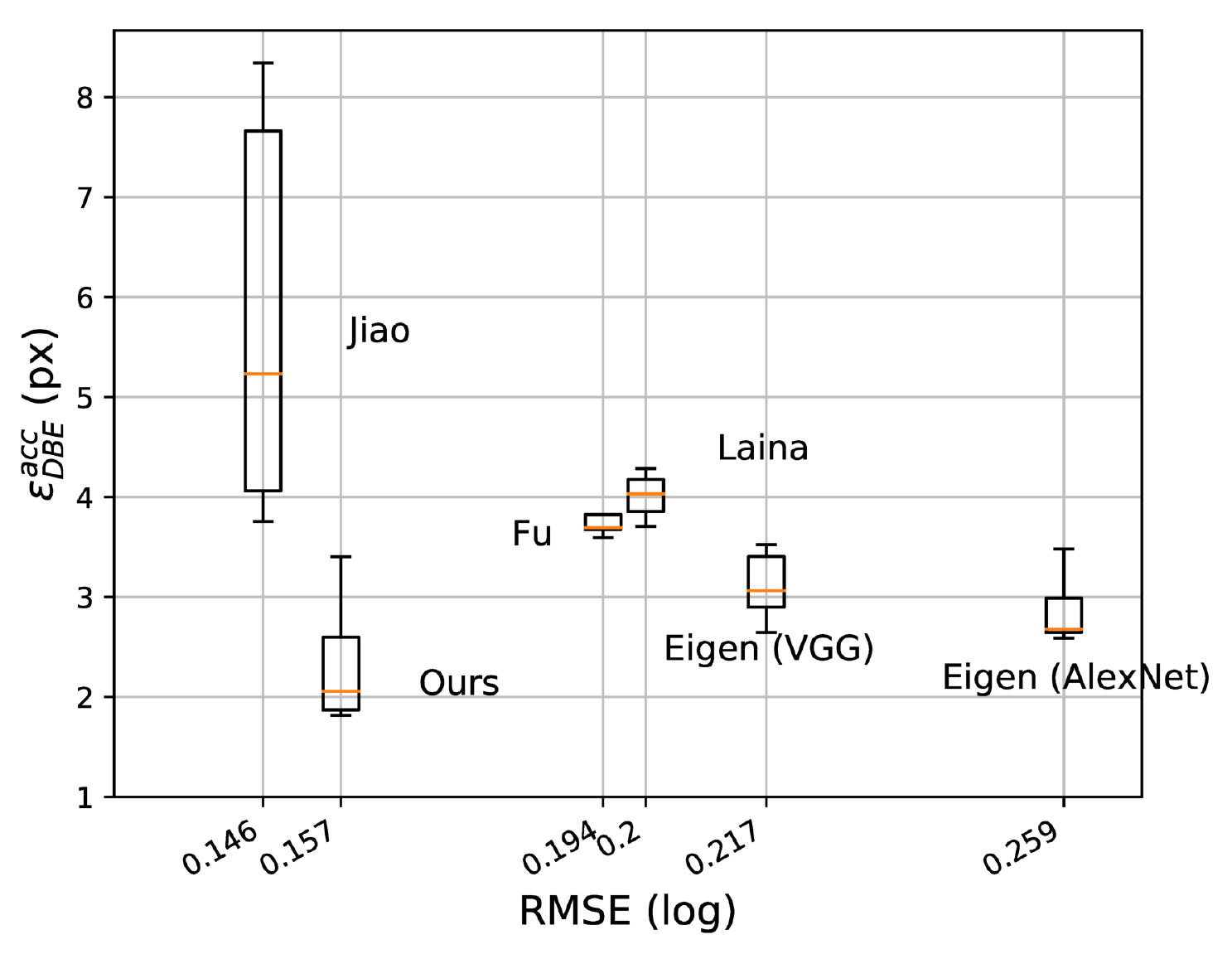}
	\end{center}
	\caption{
	Our method outperforms state-of-the-art in terms of trade-off between 
	global depth reconstruction error and occluding boundary accuracy.
	} 
	\label{fig:compare_trade_off}
\end{figure}

Since the detected edges in $\boldsymbol{\widehat{Y}}$ are highly 
sensitive to the edge detector's parameters (see Fig.\ref{fig:multi_canny}), 
we evaluate the DBE accuracy $\epsilon_{DBE}^{acc}$ using many 
random combinations of threshold parameters $\sigma^{+}$ and $\sigma^{-}$ of the
Canny edge detector. The results are shown in Fig.~\ref{fig:compare_trade_off}.

\subsection{Ablation Study}

To prove the impact of our geometry consensus terms, we performed an ablation 
study to analyze the contribution of training with synthetic and real data, as 
well as our novel geometry consensus terms. Evaluation of different models on 
our NYUv2-OC dataset are shown in Table~\ref{tab:ablation_study}, confirming 
their contribution to both improved depth reconstruction results over the whole 
NYUv2-Depth dataset and occluding contours accuracy. 

\begin{table}[t]
	\tiny
	\begin{center}
%
\begin{tabular}{|@{\hspace{0.2em}}c@{\hspace{0.2em}}|@{\hspace{0.2em}}c@{}
		|@{\hspace{0.1em}}c@{}|
		@{\hspace{-0.1em}}c@{}
		@{\hspace{-0.1em}}c@{}
		@{\hspace{-0.1em}}c@{}
		@{\hspace{-0.1em}}c@{}|}
				\hline					
				Method & Training Dataset & $RMSE_{log}$ &
				\multicolumn{4}{c|}{$\epsilon_{DBE}^{acc}$ 
					$\downarrow$(px) $\{\sigma^{-}, \sigma^{+}\}$}\\
				~ & ~ & ~ & $\{0.1, 0.2\}$ & $\{0.01, 0.1\}$ & $\{0.005, 
				0.06\}$ & $\{0.03, 
				0.05\}$ 
				\\ \hline
				w/o consensus & PBRS & 0.304$^*$ &
				2.321 & 2.751$^*$ & 3.298$^*$ & 3.380$^*$ 
				\\
				w/ consensus & PBRS & 0.262 &
				\textbf{2.046} & \textbf{2.332} & \textbf{2.574	} & 
				\textbf{2.645}\\
				w/o consensus & PBRS + NYUv2 & \ul{0.163} & 2.600$^*$ & 2.638 & 
				3.127 & 
				3.182 \\
				w/ consensus & PBRS + NYUv2 & \textbf{0.157} &
				\ul{2.272} &  \ul{2.629} & \ul{3.066} & \ul{3.152} \\
				\hline
		\end{tabular}
	\end{center}	
	\caption{
	Our added geometry consensus terms brings a 
		significant performance boost by guiding the depth towards learning 
		accurate occluding contours and it also helps keeping a good 
		trade-off between occluding contours accuracy and depth 
		reconstruction during the necessary fine-tuning on real RGB-D data. 
		$RMSE_{log}$ is computed over the full NYUv2-Depth dataset. Notations of Table.~\ref{tab:final_evaluation_table} are used here.}
	\label{tab:ablation_study}
\end{table}

\section{Conclusion}

In this paper, we  show that our SharpNet method is  able to achieve competitive
depth  reconstruction from  a  single  RGB image  with  particular attention  to
occluding  contours  thanks  to   geometry  consensus  terms  introduced  during
multi-task  training.   Our high-quality  depth  estimation which yields high  
accuracy occluding  contours reconstruction allows for realistic integration  
of virtual objects in real-time augmented reality as we achieve 150~fps 
inference speed.
We show  the superiority of our SharpNet over state-of-the-art by introducing a
first version  of our new NYUv2-OC occluding contours dataset, which we plan to
extend in future work. As by-products of our approach, high-quality normals and 
contours predictions can also be a useful representation for other computer 
vision tasks.
\vspace{20pt}

{\small
\bibliographystyle{ieee}
\bibliography{string,biblio}
}

\end{document}